\documentclass[acmsmall]{acmart}
\AtBeginDocument{%
  \providecommand\BibTeX{{%
    \normalfont B\kern-0.5em{\scshape i\kern-0.25em b}\kern-0.8em\TeX}}}

\setcopyright{acmlicensed}
\copyrightyear{2023}
\acmYear{2023}
\acmDOI{XXXXXXX.XXXXXXX}

\acmJournal{JACM}
\acmVolume{37}
\acmNumber{4}
\acmArticle{111}
\acmMonth{8}

\usepackage{amsmath,amsfonts}
\usepackage{algorithmic}
\usepackage{graphicx}
\usepackage{textcomp}
\usepackage{xcolor}
\usepackage{adjustbox}
\usepackage{hyperref}
\usepackage{caption}
\usepackage{subcaption}
\usepackage{tabularx} 

\begin{document}

\title{BeyondPixels: A Comprehensive Review of the Evolution of Neural Radiance Fields}

\author{AKM Shahariar Azad Rabby}
\email{arabby@uab.edu}
\orcid{0000-0003-3994-3105}
\author{Chengcui Zhang}
\email{0000-0002-5868-6450}
\affiliation{%
  \institution{The University of Alabama at Birmingham}
  \streetaddress{Dept. of Computer Science}
  \city{Birmingham}
  \state{Alabama}
  \country{USA}
  \postcode{35294}
}

\renewcommand{\shortauthors}{Rabby et al.}

\begin{abstract}
Neural rendering combines ideas from classical computer graphics and machine learning to synthesize images from real-world observations. NeRF, short for Neural Radiance Fields, is a recent innovation that uses AI algorithms to create 3D objects from 2D images. By leveraging an interpolation approach, NeRF can produce new 3D reconstructed views of complicated scenes. Rather than directly restoring the whole 3D scene geometry, NeRF generates a volumetric representation called a ``radiance field,'' which is capable of creating color and density for every point within the relevant 3D space. The broad appeal and popularity of NeRF makes it imperative to comprehensively examine the existing research on the topic. While previous surveys on 3D rendering have primarily focused on traditional computer vision-based or deep learning-based approaches, only a handful of them discuss the potential of NeRF. However, such surveys have predominantly focused on NeRF's early contributions and have not explored its recent advances. NeRF is a relatively new technique that is continuously being investigated for its capabilities and limitations. This survey reviews recent advances in NeRF, categorizes, and compares and contrasts them according to their architectural designs, especially in the field of novel view synthesis.
\end{abstract}

\ccsdesc[500]{Computing methodologies~Neural networks}

\keywords{Neural Radiance Field, NeRF, Computer Vision Survey, Novel View Synthesis, Neural Rendering, 3D Reconstruction, Differentiable Rendering}


\maketitle

\section{Introduction}
Image-based view synthesis techniques are widely applied to computer graphics and computer vision. One of the pressing concerns of these techniques is representing a 3D model or scene using the information of the input 2D images. Neural Radiance Field or NeRF \cite{1}, is a novel technique that trains AI algorithms to generate 3D objects from 2D images. NeRF can render novel 3D reconstructed views of complex scenes utilizing an interpolation approach between scenes. However, instead of directly recovering the entire 3D scene geometry, NeRF computes a ``radiance field'', a volumetric representation that generates color and density for each point in the concerned 3D space. 

NeRF uses a deep neural network to represent complex scenes in a fully connected, non-convolutional way. The input to the network is a continuous 5D coordinate, while the output is the volume density and view-dependent emitted radiance at that particular spatial location. 5D refers to a five-dimensional space representing 3D scenes or objects. This space includes three dimensions for spatial coordinates $(x, y, z)$ and two dimensions for camera viewing angles $(\theta, \phi)$. The method uses classic volume rendering techniques to synthesize views and is optimized using a set of images with known camera poses. The method outperforms previous work on neural rendering and view synthesis and can render novel photorealistic views of scenes with complicated geometry and appearance. The inventors of NeRF propose a new method for view synthesis by directly optimizing the parameters of a continuous 5D scene representation. They introduce a positional encoding and a hierarchical sampling strategy to address the inefficiency of the basic implementation. Using input viewing direction, their method can represent non-Lambertian effects such as specularities. NeRF uses a separate neural continuous volume representation network for each scene. In \cite{1}, the authors demonstrate NeRF's superior performance compared to prior works and also discuss how the work improves upon previous approaches that use multilayer perceptrons (MLP) to represent objects and scenes as continuous functions and the potential for more progress in efficiently optimizing and rendering NeRF.

NeRF has several advantages compared to previous approaches, including:

\begin{itemize}
    \item High-quality 3D reconstructions: NeRF can create high-quality 3D reconstructions of complex scenes, including fine surface details and reflections.
    \item Improved view synthesis capabilities: NeRF can synthesize novel views of a scene from a small number of input images, allowing for virtual walkthroughs of a scene from any viewpoint.
    \item Queryable continuous: NeRF provides a continuous representation of a scene that can be efficiently queried at any point, enabling applications such as object manipulation and rendering.
    \item Unsupervised training: NeRF can be trained unsupervised, meaning it can learn to reconstruct a scene without explicit supervision.
    \item Wide applicability: NeRF can be applied to a wide spectrum of scenarios, including outdoor scenes, indoor scenes, and even microscopic structures.
\end{itemize}

A generous amount of research based on 3D scene representation has further enhanced and extended NeRF and achieved satisfactory accuracy along with efficiency \cite{EfficientNeRF}. This new and emerging field of research can be divided into two broad categories based on its specific purpose: the analysis and improvement of NeRF itself \cite{3,4,5} and the extensions based on the NeRF framework \cite{star, NeRD, GIRAFFE}. Over time, researchers have developed various innovative techniques to overcome the limitations of NeRF, described in more detail as follows.

\begin{figure}[htbp]
\begin{center}    \includegraphics[width=.48\textwidth]{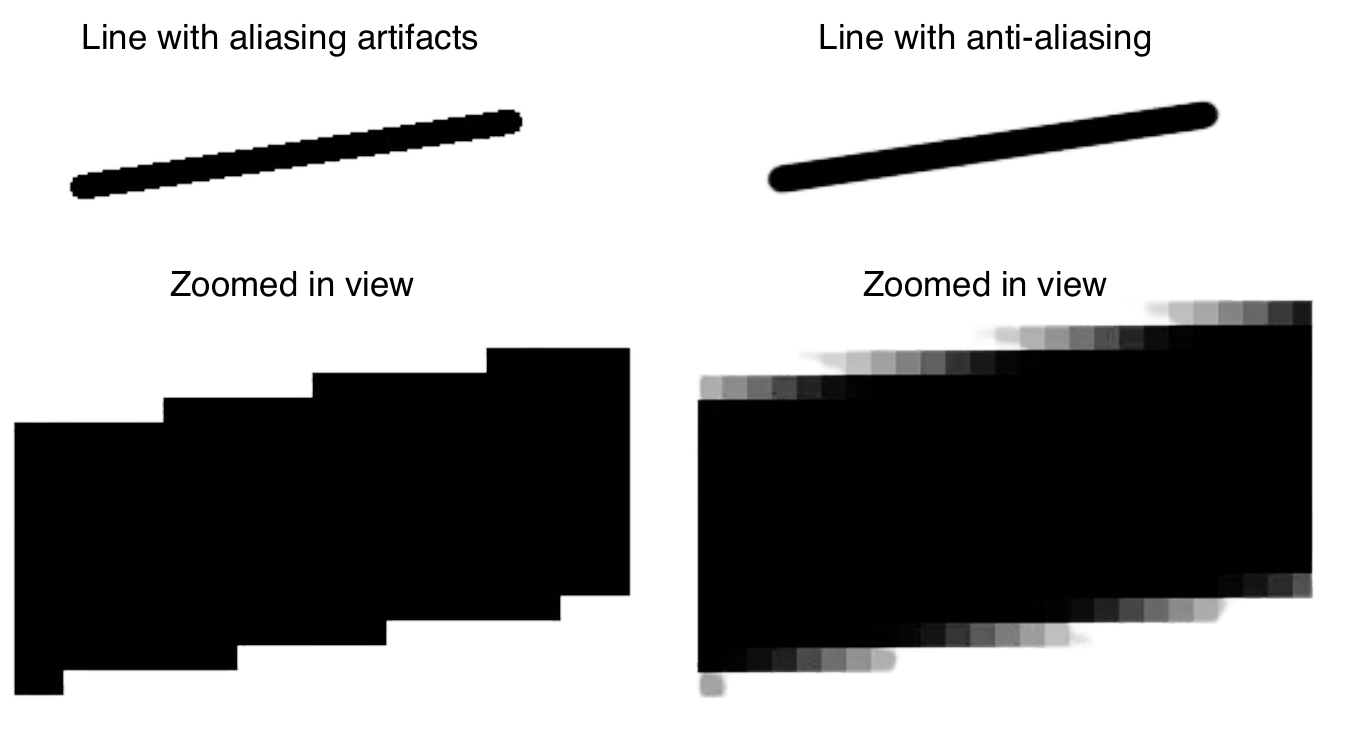}
\caption{Example of aliasing artifacts}
\label{fig:1a}
\end{center}
\end{figure}

\textbf{Rendering Quality and Scalability:} The representation of NeRF suffers from vulnerability to sampling and aliasing problems, which can lead to significant artifacts in the synthesized images.  Aliasing artifacts are the visual artifacts that appear when NeRF is used to render images of scenes that contain sharp edges or textures. Figure \ref{fig:1a} shows an example of this problem. These artifacts are caused by the limited sampling of the radiance field, which leads to the inaccurate reconstruction of these features. Researchers tried to overcome this issue by using a cone instead of rays and to avoid ray sampling in the empty scene space \cite{mipNerf, DBLP:journals/corr/abs-2111-13679, DBLP:journals/corr/abs-2112-03907}. One of the biggest challenges with NeRF is its slow training speed. Training a NeRF model requires a large amount of memory and computing power. NeRF is also slow in rendering, particularly at high resolutions. This can make it impractical for real-time applications or generating animations with multiple frames. Many researchers have focused on methods such as dividing the scene into smaller manageable blocks, voxel-based representations, and super-sampling, etc., to accelerate NeRF \cite{muller2022instant,DBLP:journals/corr/abs-2103-14024, Plenoxels, SunSC22}.

\textbf{High-Quality Large Dataset:} One limitation of NeRF is that it requires a large dataset of high-quality images to train the network. This dataset needs to capture the scene from various viewpoints, which can be challenging and time-consuming. The dataset's quality also plays a critical role in the network's performance. If the dataset is noisy or contains artifacts, it can significantly impact the quality of synthesized views. Furthermore, as the scene's complexity increases, the dataset demands also increase, making it challenging to use NeRF for large-scale scenes. Researchers tried to focus on these limitations and improve the performance of NeRF by adding prior conditions to the model. Prior can play a dual role in image processing. Prior can be used to assist in the reconstruction of images or can be employed in a generative fashion to produce new images, with prior as an additional input feature or a label used to guide the model's predictions. For example, in image generation, a conditional deep-learning model might be trained to generate images of specific objects or scenes based on a given set of prior such as an object class or a scene description \cite{MedNeRF, DBLP:journals/corr/abs-2103-15606, kosiorek2021nerf, NeuForm}. Furthermore, some researchers also tried a one-few-zero shot approach to address these limitations. These different forms of learning allow NeRF to generate accurate 3D models with limited or no data, making them suitable for real-world applications \cite{yu2021pixelnerf, PANeRF, AutoRF}.
  
\textbf{Representing Articulated Objects:} NeRF has shown remarkable progress in representing complex and highly detailed static scenes using deep learning models. However, one of the significant limitations of NeRF is its inability to effectively represent articulated objects, such as humans or animals with multiple moving parts, where the object’s geometry is highly dynamic and can change with respect to the viewpoint. The problem arises due to the fact that NeRF represents the scene as a continuous volumetric function and assumes that the scene is static. When applied to articulated objects, this assumption no longer holds, resulting in artifacts in the synthesized images. In such cases, the explicit modeling of each object part using a set of parametric functions is not feasible, and hence, the representation of these objects requires new approaches. Recent works have proposed approaches to handle articulated objects by learning a deformation field to map from a canonical space to the current frame \cite{pumarola2021d}. This allows modeling object motions while still exploiting temporal redundancy through the shared canonical configuration. However, these methods have only been demonstrated for relatively simple articulated motions. More complex articulated objects such as humans remain challenging for NeRF-based approaches. A trending application of NeRF is for articulated models of people or cats. Such models represent the shape of the object in the image using rigid parts that are interconnected \cite{iNeRF, DBLP:journals/corr/abs-2104-06405, NeRF-Pose, humannerf2}. 

\textbf{Scene Editing:} One other limitation of NeRF is the difficulty in editing the scene after it has been rendered. The reason is that NeRF represents the scene as a continuous function and does not store explicit geometry information. This makes it challenging to modify or remove objects in the scene or to change their properties, such as texture or lighting. While some recent works have attempted to address this limitation by proposing methods for interactive scene editing, these works are still in their infancy and will require significant improvements to become practical tools for artists and designers \cite{cagenerf, fenerf, conerf, NeuPhysics}. 

NeRF has also been applied in several other related fields with excellent results \cite{RL, SNAKE, xrnerf}, including the field of audio, where it has found use due to the similarity of sound propagation to rays in a volume \cite{arxiv.2204.00628, suinras}.

While there have been several surveys and research papers discussing the traditional computer vision-based \cite{26, 27, 28, 29} and conventional deep learning-based approaches \cite{30, 31, 32, 33} to 3D rendering, only a few of them have discussed NeRF \cite{chen2021towards}. This is because NeRF is a relatively new technique that has only been introduced recently, and its full potential and limitations are still being explored. Traditional computer vision and deep learning approaches have been limited to generating photorealistic images with intricate geometries using discrete representations such as triangle meshes or voxel grids. In contrast, NeRF has shown remarkable potential in generating highly realistic images with precise color and geometric position. By tracing camera rays through a scene, NeRF obtains a set of sampled 3D points, which are utilized alongside their corresponding 2D viewing directions as input to a neural network. The output set comprises colors and densities, which collectively generate visually accurate scene representations.
Also, NeRF requires fewer images to generate a 3D scene than traditional computer vision and deep learning approaches, reducing the rendering time. Researchers in traditional computer vision societies focused on mathematical solutions such as Marching Cubes and Space partitioning, while Marching Cubes \cite{lorensen1987marching, nugroho20163d, ccalicskan2017three} seems the most popular algorithm for surface rendering. Some researchers have also used other methods such as Contour Filter \cite{hafizah2010development} and various interpolation methods \cite{ghoshal20203d}. In deep learning, during the early times, CNN was used by several for surface rendering, while for volume rendering, traditional computer vision algorithms were still used \cite{li20193d,tong2020x}. Later, new traditional computer vision-based algorithms have been used for volume rendering, such as space partitioning, occupancy networks \cite{mescheder2019occupancy}, shape partitioning, and subspace parameterization \cite{ riegler2017octnet}.

In this survey,  After outlining the detailed explanation of the theory behind NeRF in Section 2, we categorize and analyze the existing research works from Sections 3 to 7 based on the abovementioned classification. We discuss the strengths and limitations of NeRF-based methods, including their advantages in terms of representation capabilities, rendering quality, and generalization to new scenes, as well as challenges in terms of interpretability, computational efficiency, and scalability in Section 8. Finally, we highlight the potential future research directions of NeRF-based methods, such as improving the interpretability, efficiency, and scalability of the method.

\section{Background}
NeRF represents the geometry and appearance of a scene as a continuous, volumetric radiance field parameterized by a neural network. The rendering process involves querying the neural network to predict the color and density values along camera rays, which are then composed into a final image using volume rendering techniques. NeRF's ability to synthesize high-quality images from arbitrary viewpoints has revolutionized novel view synthesis in computer graphics and has found applications in areas such as virtual and augmented reality, video games, and movie production. Various datasets, loss functions, assessment metrics, and literature reviews are discussed to provide a comprehensive understanding of NeRF's capabilities and advancements in the field. 

\begin{figure}[htbp]
\begin{center}    \includegraphics[width=\textwidth]{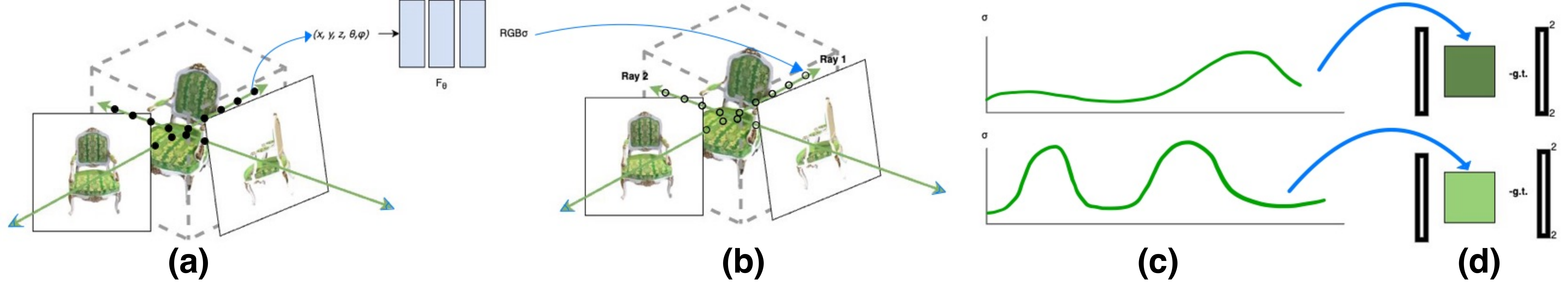}
\caption{An overview of the process of rendering and training NeRF. The first step (a) involves selecting sampling points for each pixel in an image that needs to be synthesized. The next step (b) involves using NeRF's MLP(s) to generate densities and colors at the selected sampling points, followed by volume rendering techniques to composite these values into an image (c). Since this rendering function is differentiable, scene representation is optimized by minimizing the difference between the synthesized and observed ground truth images (d).}
\label{fig:1}
\end{center}
\end{figure}
\subsection{Understating NeRF}
Neural Radiance Fields (NeRF) \cite{1} is a groundbreaking technique in computer graphics that allows for photorealistic rendering of 3D scenes. Developed in 2020 by researchers at the University of California, Berkeley, NeRF uses deep neural networks to model the 3D geometry and appearance of objects in a scene, creating high-quality visualizations that are difficult or impossible to achieve using traditional rendering techniques. 

The key idea behind NeRF is to represent the appearance of a scene as a function of 3D position and viewing direction, known as the radiance field. The radiance field describes how light travels through the scene and interacts with its surfaces and can be used to generate images from arbitrary viewpoints. The ``neural'' part of the name refers to the fact that the radiance field is learned using a neural network. The NeRF algorithm involves several steps: data acquisition, network training, and rendering. Figure \ref{fig:1} provides an overview of the NeRF scene representation and differentiable rendering procedure. It shows the steps involved in synthesizing images, which include sampling 5D coordinates along camera rays, using an MLP to generate color and volume density, and compositing these values into an image using volume rendering techniques. 

In the following, we will explore the basics of NeRF, how it works, and its applications in computer vision and computer graphics.

\subsubsection*{Basics of NeRF}
NeRF is a deep learning technique that learns to model a scene's 3D geometry and appearance by capturing the relationship between the scene's geometry and appearance using a neural network. The neural network takes a set of 2D images of the scene from different viewpoints as input and outputs a 3D representation of the scene. This representation can render new views of the scene from any viewpoint, resulting in highly photorealistic renderings.

\subsubsection*{Data Acquisition}

The first step in the NeRF algorithm is to acquire a set of 2D images of the scene from different viewpoints. Next, these images are used to train the neural network to model the relationship between the 3D geometry and the scene's appearance. Ideally, the images should cover a wide range of viewpoints and lighting conditions to ensure the neural network can capture the full range of variability in the scene.

\subsubsection*{Network Training}

The next step in the NeRF algorithm is to train a neural network to model the relationship between the 3D geometry and the scene's appearance. Specifically, the neural network takes as input a pair of coordinates $(x, y)$ that corresponds to a 2D point in the image plane and a viewing direction vector and outputs a corresponding 3D point in the scene and its associated color. This process is repeated for every pixel in the input images, resulting in a set of 3D points and colors that can be used to render the scene from any viewpoint.

The training process involves optimizing the parameters of the neural network to minimize the difference between the rendered image and the ground truth image. This is done by computing the rendering equation, which describes how light travels through a scene and interacts with its surfaces, and using it to generate a synthetic image that is compared to the ground truth image.

\subsubsection*{Rendering}
The final step in the NeRF algorithm is to use the trained neural network to render new views of the scene from arbitrary viewpoints. This is done by querying the neural network for the radiance field at each point in the 3D space and using it to generate an image of the scene from the desired viewpoint.

\subsubsection*{Applications}
NeRF has already been used in various applications, and its potential for future applications is vast. For example, NeRF could generate realistic 3D models of archaeological sites, allowing researchers to explore and study them in new ways. It could also generate virtual try-on experiences for online shopping, allowing customers to see how clothes would look on them before making a purchase.
NeRF has a wide range of applications in computer graphics, including virtual reality, augmented reality, video games, and movie production. By enabling the creation of highly realistic 3D visualizations, NeRF has the potential to revolutionize how we create and interact with digital content.

One of the most promising applications of NeRF is in virtual and augmented reality. NeRF can create immersive experiences indistinguishable from reality by generating real-time photorealistic renderings of virtual environments. This has the potential to transform how we interact with virtual content, making it possible to create realistic simulations of real-world scenarios for training, education, and entertainment.

Another application of NeRF is in video game development. Using NeRF to create highly realistic environments and characters, game developers can create more immersive and engaging experiences for players. This can lead to a more enjoyable and satisfying gaming experience, which can, in turn, drive increased engagement and revenue for game developers.

In movie production, NeRF can be used to create highly realistic special effects and CGI scenes. By modeling the appearance and geometry of objects and scenes using neural networks, filmmakers can create highly detailed and realistic visualizations that are difficult or impossible to achieve using traditional rendering techniques.

\subsection{Datasets for NeRF}
The performance of Neural Radiance Fields (NeRF) models depends heavily on the dataset used for training and evaluation. Choosing appropriate datasets is crucial to training and assessing the capabilities and limitations of NeRF methods under different conditions. This section discusses key datasets that have been pivotal for NeRF research.

\subsubsection{Synthetic Datasets}
Synthetic datasets generated using computer graphics engines offer great value for training and analyzing NeRF models. They provide full control over scene properties and ground truth information that is challenging to obtain from real datasets.

\textbf{Diffuse Synthetic 360°:}
The Diffuse Synthetic 360° \cite{1} dataset contains renderings of 4 simple Lambertian objects with low geometric complexity: chair, pedestal, cube, and vase. The objects were scanned using a 3D scanner and texture mapped, then rendered at 512x512 pixel resolution using a graphics engine. The dataset includes 479 input views that are sampled evenly across the upper hemisphere surrounding each object. An additional 1,000 views are rendered, covering the full viewing sphere, and used as held-out test data. Since the objects have perfectly diffuse (Lambertian) reflectance, the rendered images have lighting variation but no specularities or viewpoint-dependent effects. The inputs to the dataset are simply RGB images from the known viewpoints. The outputs are also RGB images rendered from novel viewpoints sampled across the sphere. This dataset offers a simple scenario for testing view synthesis techniques, with multiple views of objects that have flat shading and minimal geometric complexity. A strength is the dense sampling of hundreds of views covering the entire upper hemisphere for each object. 



\textbf{Realistic Synthetic 360°:}
The Realistic Synthetic 360° dataset \cite{1} contains renderings of 8 objects with complex geometry and realistic non-Lambertian materials: chair, drums, ficus, hotdog, lego, microphone, and ship. The objects were modeled in Blender \cite{blender_soft} and rendered using its Cycles path tracer, exhibiting effects such as specularities, translucency, interreflections, and cast shadows. Each object was rendered at 800x800 pixel resolution from 100 randomly sampled viewpoints across the upper hemisphere, providing the input views. An additional 200 views across the full sphere were rendered and held out for testing. Because the inputs come from a physics-based path tracer, the dataset contains photorealistic effects missing from the simpler Lambertian dataset. The inputs are RGB images rendered from the known viewpoints. The outputs are also RGB images rendered from the sphere's novel views. This dataset offers a more challenging test case for novel view synthesis, requiring generalization to new object materials and lighting.

\textbf{Real Forward-Facing dataset:}
The Real Forward-Facing dataset \cite{mildenhall2019local} comprises eight real-world scenes captured using a handheld cellphone camera. The scenes are room, fern, leaves, fortress, orchids, flowers, t-rex, and horns. Each scene was captured by taking 20 to 62 images with the camera pointed roughly forward, resulting in standard perspective viewpoints. The images were captured at 1008x756 pixel resolution. Since the images come from casual smartphone photography, the dataset exhibits challenges such as motion blur and under/overexposure. For each scene, 1/8 of the images were held out for testing. Camera intrinsics and poses were estimated using the COLMAP structure-from-motion system [39]. The inputs are simply the RGB smartphone photos and the estimated camera parameters. The outputs are novel views rendered from the optimized neural radiance field model. This dataset offers a practical test case for novel view synthesis using real-world photos with challenging uncontrolled capture conditions. A limitation is the relatively sparse forward-facing view sampling.

\textbf{NSVF Synthetic Dataset:} NSVF Synthetic Dataset \cite{liu2020neural} is created for Neural Sparse Voxel Fields (NSVF) research. This Synthetic-NSVF dataset contains eight synthetic objects rendered at 800x800 resolution with complex geometry and lighting effects. The objects showcase diverse 3D models, including a wine holder, steam train, toad, robot, bike, palace, spaceship. Each object is rendered from 100 views for training and 200 views for testing, with the input being posed RGB images and the output a novel rendered view. The authors created this dataset to complement the existing synthetic-NeRF dataset with more complex and diverse scenes to evaluate the NSVF method. The images exhibit challenging materials, geometries, and lighting that push the boundaries of novel view synthesis techniques. As a synthetic dataset, it provides perfect ground truth novel views for quantitative evaluation. The Synthetic-NSVF dataset provides a challenging benchmark for evaluating representation learning and novel view synthesis methods on complex 3D scenes.

\textbf{BlendedMVS Dataset:} The BlendedMVS dataset \cite{yao2020blendedmvs} is a large-scale synthetic dataset introduced for training multi-view stereo networks. It contains over 17,000 high-resolution images covering 113 scenes, including cities, architecture, sculptures, and small objects.
The dataset generation pipeline starts with reconstructing textured 3D meshes from input images using an online 3D reconstruction service. The textured meshes are then rendered from different viewpoints to generate corresponding color images and depth maps. Using a frequency domain filtering approach, the rendered color images are blended with the original input images to introduce realistic lighting effects. Specifically, high-frequency details are extracted from the rendered images while low-frequency lighting effects are extracted from the original images and then fused to create the final blended training images.
The blended images retain detailed surface textures for consistency with the rendered depth maps while also preserving realistic environmental lighting from the original images. This helps models trained on BlendedMVS generalize better to real-world datasets than training only on rendered or input images.
The dataset contains 113 scenes with 20-1000 images per scene. The images and depth maps are provided at a unified resolution of 1536x2048 pixels. Camera parameters are also included. Online data augmentation with random brightness, contrast, and blur is employed during training.

\subsubsection{Real Scene Datasets}
Real-scene datasets, often reconstructed from photographs or scans, provide essential benchmarks for analyzing NeRF algorithms' performance on real-world data.

\textbf{DTU Multi-View:} The DTU Multi-View Stereo dataset \cite{jensen2014large} contains a set of 80 real-world scenes with image sequences and structured light 3D scans as reference data. The scenes contain a variety of objects and materials, including houses, toys, groceries, shiny objects, and more. The dataset was captured by placing objects on a turntable and capturing images from 49 or 64 precisely calibrated camera viewpoints per scene. The cameras are positioned at two heights, 50cm and 65cm from the object. The reference 3D scans were captured using a structured light technique from each camera viewpoint and fused into a single high-resolution point cloud per scene. 
The dataset provides the Image sequences (49 or 64 images per scene) in uncompressed PNG format with a resolution of 1600x1200 pixels. The images have uniform directional lighting and camera calibration information, including intrinsic parameters and extrinsic pose for each viewpoint. It also provides fused high-resolution 3D point clouds from the structured light scans in XYZ text format with 10-14 million points per scene point cloud. The points are roughly sampled at 0.2mm resolution, and observability masks indicate where the reference 3D points have data.

\textbf{Matterport3D:} Expanding on DTU, the Matterport3D dataset \cite{chang2017matterport3d} contains RGB-D data capturing 90 indoor scenes, comprising 10,800 panoramic views from 194,400 RGB-D images. The scenes cover over 219,000 square meters of surface area, including homes, offices, churches, and more. A Matterport camera captured each scene on a tripod rotated to 18 positions at each of 10,800 locations to generate panoramic skybox images. The dataset provides HDR RGB images at 1280x1024 resolution, depth images, camera intrinsics and extrinsic, textured 3D mesh reconstructions, and 2D and 3D semantic annotations labeling regions, objects, and voxels. Key characteristics are the large scale (90 buildings), accurate global alignment, and comprehensive set of viewpoints, sampling each scene surface from around 11 viewpoints on average. The color images, depths, camera poses, textured meshes, and semantic annotations support tasks such as novel view synthesis, 3D reconstruction, segmentation, and scene understanding. Overall, Matterport3D is a large-scale, versatile dataset enabling research on learning representations of indoor scenes from images and 3D data.

\textbf{DeepVoxels-Real Dataset:} Comprising five tabletop real scenes imaged by 352 cameras in a dome, this dataset \cite{sitzmann2019deepvoxels} captures complex real geometries and materials. Ground truth meshes and semantics are included, offering an excellent testbed for NeRF on real small-scale scenes.

While synthetic datasets remain widely used for developing NeRF techniques, real datasets are critical for evaluating performance in real-world conditions. Datasets spanning indoor/outdoor scenes, objects, and forward/360° capture will continue enabling NeRF advances and measuring progress in challenging yet practical scenarios. Carefully curating datasets with ground truth will further model analysis and development.

\subsection{Loss Function}
NeRF has revolutionized novel view synthesis of complex 3D scenes by representing them as a continuous volumetric function that maps 5D coordinates of rays to properties such as color and density. This approach renders high-quality novel views by querying the implicit neural representation. NeRF learns these continuous scene representations from only sparse input views. The loss function is crucial in optimizing NeRF models by measuring the difference between the rendered outputs and ground truth pixel data from the input views. It guides the learning process by providing a training signal to minimize the error between the model's rendered results and the target views. The loss function optimizes the continuous volumetric neural representation to reproduce the view-dependent effects observed in the input images accurately. Choosing the proper loss function is critical for enabling high-fidelity view synthesis while also imposing the desired priors and inductive biases. Recent works have explored losses beyond view consistency, incorporating objectives to support neural animation, disentanglement of dynamic/static elements, enhanced realism through adversarial training, and selective editing abilities. Loss functions are a core component of NeRF optimization that determines the capabilities and quality of the learned volumetric scene representations.

The most common and essential loss is a photometric loss between rendered and ground truth pixel colors. This serves as the primary objective for optimizing view consistency \cite{tewari2022advances}. The original NeRF paper defines a simple RGB reconstruction loss $L_c$ to match rendered color $\hat{C}_r$ and ground truth $C_r$ per ray $r$ \cite{1}:

\begin{equation}
L_c = \sum_{r \in R} ||C_r - \hat{C}_r||_2^2
\label{eq:loss1}
\end{equation}

In Equation \ref{eq:loss1}, $R$ is the set of sampled camera rays and this $L_2$ loss supervises the implicit neural representation to reproduce effects such as view-dependent lighting and materials seen in the input views, the key to optimizing novel view realism.

While the photometric loss focuses on view synthesis quality, additional losses are incorporated to enable specific capabilities. For animating scenes, Chen et al. use a pose regularization loss $L_p$ to prevent optimized poses from deviating too far from initial estimates \cite{Animatable}:

\begin{equation}
L_p = \lambda_1||\theta_t - \hat{\theta}_t|| + \lambda_2||\theta_t - \hat{\theta}_{t-1}||
\label{eq:loss2}
\end{equation}

In Equation \ref{eq:loss2}, $\theta_t$ is the initial pose parameters, $\tilde{\theta}_t$ and $\tilde{\theta}_{t+1}$ are the optimized pose parameters of frames $t$ and $t+1$. $\lambda_1$ and $\lambda_2$ are the corresponding penalty weights. The pose regularization loss $L_p$ does depend on $t$, since it is computed for each frame $t$. 

Kobayashi et al. employ an image feature loss $L_f$ to preserve semantic content better \cite{dff}:
\begin{equation}
L_f = \sum_{r \in R} \left\| \hat{F}(r) - f_{\text{img}}(I, r) \right\|_1 
\label{eq:loss3}
\end{equation}

In Equation \ref{eq:loss3}, \(L_f\) represents the feature loss and quantifies the \(L_1\) disparity between the rendered features \(\hat{F}(r)\), obtained through volume rendering of the feature field \(f\), and the corresponding features \(f_{\text{img}}(I,r)\) produced by the teacher network for ray \(r\) within image \(I\). The feature loss is employed in conjunction with the photometric loss \(L_p\) to train the complete distilled feature field model effectively.

Some methods incorporate losses to disentangle scene components such as static and dynamic elements. Wu et al. use an binary entropy loss $L_b$ to decompose them \cite{D2NeRF}:

\begin{equation}
L_{b}(r, \tau_{i}) = -\int_{t_{n}}^{t_{f}} H_{b}(w(r(t), \tau_{i})) dt
\label{eq:loss4}
\end{equation}

In Equation \ref{eq:loss4}, \(r\) represents the camera ray, \(\tau_i\) denotes the per-frame time embedding for frame \(i\), and \(t_n\) and \(t_f\) define the near and far bounds for volumetric integration along each ray. The spatial ratio \(w(r(t))\) captures the relationship between dynamic and static density along the camera ray. The binary entropy loss \(L_b\) operates on \(w\) to penalize deviations from a binary distribution, promoting values close to 0 or 1. This constraint enforces a clear separation between dynamic and static components, reducing overlap in radiance fields. Additionally, \(L_b\) acts as a regularizer to prevent the more expressive dynamic radiance field from inaccurately representing static regions in the scene. Density regularization losses are also common to promote sparser, concentrated geometry \cite{VoxGRAF}.

Specialized losses are required to enable the editing of NeRF scenes. Liu et al. employ a density loss $L_{dens}$ to selectively remove shapes \cite{edit}:

\begin{equation}
L_{dens} = -\sum_{r \in y_f} \sigma_r^T \cdot \log(\sigma_r)
\label{eq:loss5}
\end{equation}
In Equation \ref{eq:loss5}, $\sigma_r$ represents the predicted density for a given ray $r$, $T$ is the timestep, and $y_f$ constitutes the set of rays associated with the foreground region intended for removal, in which the strategy of minimizing entropy plays a crucial role. Essentially, this approach squeezes the distribution of density values, driving them closer to 0 or 1. This, in turn, results in a sparser distribution of density predictions, particularly for the selected foreground rays, $y_f$. The intuitive objective of this process, when aiming to remove a specific shape from the scene, is to encourage the radiance field to produce low-density values along the rays intersecting with that shape. Achieving this entails minimizing the entropy loss, effectively pushing the densities closer to 0 for the relevant rays and ``cutting out'' the undesired shape.

As a brief summary, the photometric loss is the key to optimizing view synthesis quality. Additional losses such as pose regularization, feature matching, density regularization, and adversarial losses support capabilities such as animation, disentanglement, and enhanced realism. For editing, opacity and density losses provide control over scene composition and geometry. NeRF loss functions continue evolving beyond view synthesis towards controllable generation and editing.

\subsection{Assessment Metrics}

Novel view synthesis via NeRF in the standard setting employs visual quality assessment metrics for benchmarks. These metrics aim to evaluate the quality of individual images either through full-reference (with ground truth images) or no-reference (without ground truth images) methods. The most commonly used metrics in NeRF literature are Peak Signal Noise Ratio (PSNR), Structural Similarity Index Measure (SSIM), and Learned Perceptual Image Patch Similarity (LPIPS).

\subsubsection{Peak Signal to Noise Ratio (PSNR)}
Peak Signal Noise Ratio (PSNR) is a full-reference quality assessment metric measuring the distortion between the reference and the assessed images. The calculation, described by Equation \ref{eq:1}, involves the ratio of the highest attainable pixel value to the mean squared error (MSE) between the two images. A higher PSNR value indicates a lower MSE and a better-quality image. However, PSNR is not very sensitive to subtle changes in the image, such as brightness, contrast, or saturation changes. 
\begin{equation}
    PSNR(I) = 10 \cdot \log_{10}\left(\frac{MAX(I)^2}{MSE(I)}\right)
    \label{eq:1}
\end{equation}
where $MAX(I)$ represents the maximum possible pixel value in the image, and $MSE(I)$ denotes the pixel-wise mean squared error calculated over all color channels. PSNR is widely used in NeRF research and in various signal-processing fields.

\subsubsection{Structural Similarity Index Measure (SSIM)}
Structural Similarity Index Measure (SSIM) is a full-reference quality assessment metric that measures the similarity between the reference and the assessed images in terms of their structural content. Equation \ref{eq:2} defines SSIM as the product of two components: the local similarity and the contrast similarity. The local similarity measures the similarity of the local pixel intensities between the two images, while the contrast similarity measures the similarity of their local contrasts. A higher SSIM value indicates a higher similarity between the two images and a better quality image. SSIM is more sensitive to subtle changes in the image than PSNR, but it is also more difficult to compute.
\begin{equation}
    SSIM(x, y) = \frac{(2\mu_x\mu_y + C_1)(2\sigma_{xy} + C_2)}{(\mu_x^2 + \mu_y^2 + C_1)(\sigma_x^2 + \sigma_y^2 + C_2)}
    \label{eq:2}
\end{equation}

SSIM metric quantifies the similarity between two images \(x\) and \(y\), by comparing various statistical properties of the images, including their means (\( \mu_x \) and \( \mu_y \)), variances (\( \sigma_x^2 \) and \( \sigma_y^2 \)), and the covariance (\( \sigma_{xy} \)) between them. Additionally, \(C_1\) and \(C_2\) are constants that prevent division by very small variances, ensuring numerical stability in the calculation. This equation quantitatively measures how similar two images are in terms of structure and texture.

\subsubsection{Learned Perceptual Image Patch Similarity (LPIPS)}
Learned Perceptual Image Patch Similarity (LPIPS) is a full-reference quality assessment metric that uses a deep neural network to learn the perceptual similarity between images. It is able to capture the subtle differences between images that are not captured by traditional metrics such as PSNR and SSIM. LPIPS is the most recent metric and the most sensitive to subtle changes in the image. However, it is also the most computationally expensive metric. Equation \ref{eq:3} outlines the LPIPS score derived from the weighted pixel-wise Mean Squared Error (MSE) of feature maps across multiple layers. A lower LPIPS value indicates a higher perceptual similarity between the two images and a better quality image.
\begin{equation}
    LPIPS(x, y) = \sum_{l=1}^{L} \frac{1}{H_l W_l} \sum_{h=1}^{H_l} \sum_{w=1}^{W_l} ||w_l(x_{lhw} - y_{lhw})||_2^2
    \label{eq:3}
\end{equation}

$x_{lhw}$ and $y_{lhw}$ represent the features of the reference and assessed images at the pixel at coordinates $(w, h)$ and layer $l$. $H_l$ and $W_l$ correspond to the height and width of the feature map at layer $l$.
\subsection{Literature Review}
NeRF has received significant attention in the computer vision research community in recent years, and a number of survey papers have been published to provide a comprehensive overview of the advances made in this area. In the survey paper \cite{kato2020differentiable}, the authors discuss the advantages of NeRF over other rendering techniques. The first survey paper to focus on NeRF is a pre-print by Dellaert et al. \cite{dellaert2020neural}, which references a relatively small number of publications at the time and categorizes those papers into different types. In contrast, Tewari et al. \cite{tewari2022advances} wrote a comprehensive state-of-the-art report on advances in neural rendering with a specific focus on NeRF models. This paper includes several influential NeRF papers and other cutting-edge rendering techniques. The authors provide detailed descriptions and analyses of the discussed papers, making the report a valuable resource for researchers and practitioners in this field. Gao et al. \cite{gao2022nerf} published a survey paper regarded as one of the most comprehensive surveys of NeRF. To organize the vast amount of NeRF literature, the authors classify the literature into six categories based on their citation counts and GitHub stars. In addition, the authors discuss the datasets and evaluation metrics used in various NeRF papers. While the authors have attempted to cover a large number of papers, the survey provides less detailed information on many papers, making it difficult to follow without expert knowledge. Furthermore, the survey lacks a clear technical justification for the categories, which may leave readers wondering about the rationale behind the categorization. In addition, there are websites \cite{NeRFExpl88:online, NeRFatIC66:online, NeRFatCV90:online, NeRFatNe37:online} that periodically summarize and categorize NeRF papers from top computer science conferences. However, these resources generally offer only brief summaries but not comprehensive surveys.

Our survey stands apart from previous surveys in that we focus exclusively on papers related to NeRF, providing comprehensive and detailed summaries and comparisons of recent works. To aid future researchers, we propose a new classification system for NeRF models based on the challenges those models address to encourage further research to overcome these challenges. By extensively covering the state-of-the-art NeRF-based papers presented at top computer science venues, our survey offers a comprehensive and up-to-date resource for researchers and practitioners interested in this cutting-edge technology.

\section{Rendering Quality}
The original NeRF method creates a volumetric representation of a scene using input images and camera poses, which can result in undersampling and aliasing artifacts. In addition, the training process is slow, and rendering can take a long time. Researchers have tried to overcome these issues by using cones instead of rays and implementing methods such as dividing the scene into smaller blocks or using voxel-based representations, as described below.

\begin{figure}[htbp]
\begin{center}    \includegraphics[width=0.7\textwidth]{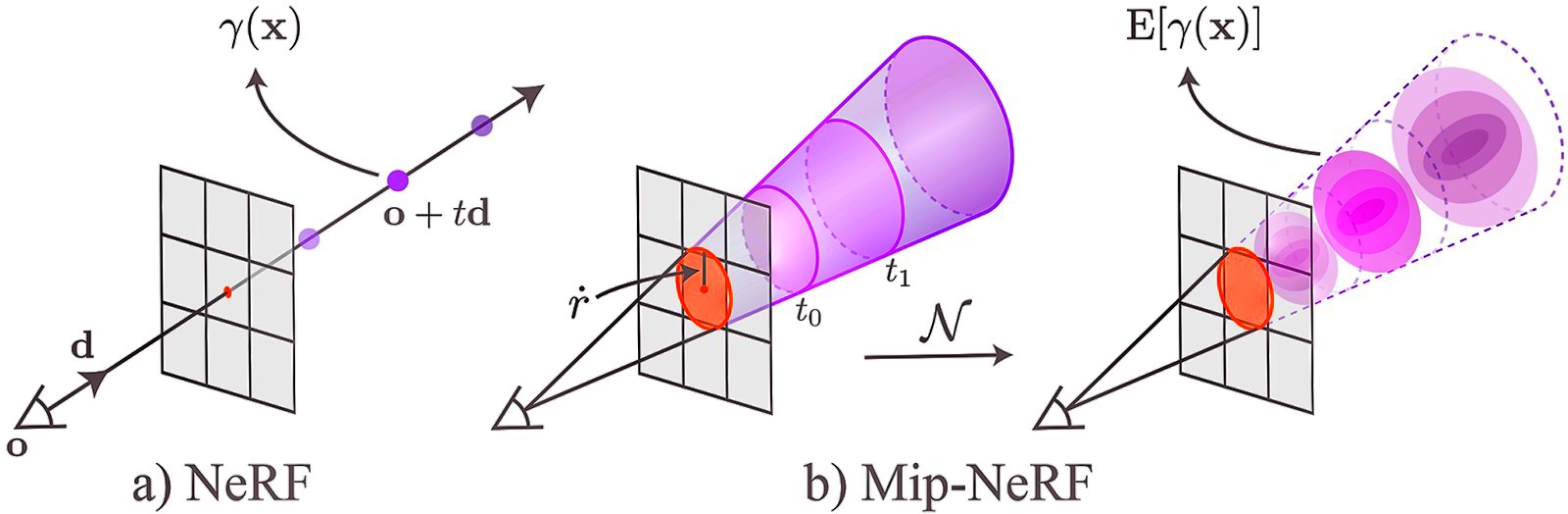}
\caption{Comparing NeRF and Mip-NeRF neural rendering methods. (a) NeRF samples point along rays traced from the camera center through each pixel. These points are encoded with a positional encoding to produce features. (b) Mip-NeRF instead reasons about the 3D conical frustum defined by a camera pixel. These frustums are featurized using an integrated positional encoding, which approximates the frustum as a multivariate Gaussian and computes the integral of the positional encodings within the Gaussian. This figure is adapted from Mip-NeRF paper \cite{mipNerf}.}
\label{fig:mipnerf}
\end{center}
\end{figure}

Mip-NeRF \cite{mipNerf} is a novel improvement of NeRF, designed to represent 3D objects as a continuous function. While NeRF was an innovative solution to the problem of rendering 3D objects with high accuracy, it was plagued by sampling and aliasing problems, which led to reduced precision in the resulting image. Mip-NeRF addresses these issues by introducing cone tracing instead of the ray tracing used in NeRF, which allows for a more accurate representation of each pixel. The cone is divided into conical frustums \cite{FrustumW66}, and a representation of the volume covered by each frustum is created using integrated positional encoding (IPE) features, which replaces NeRF's two separate ``coarse'' and ``fine'' MLPs with a single multiscale MLP, reducing the model size and making training and evaluation faster. IPE is a technique that allows neural networks to take advantage of the geometric structure of the data. In the case of Mip-NeRF, IPE features are used to encode the position and orientation of each point in 3D space, allowing the network to understand better the geometry of the scene being rendered. Mip-NeRF also uses a scale-aware structure, which enables the network to consider the scale of the objects in the scene. This is important because it ensures the network can accurately represent objects of different sizes and resolutions. The scale-aware structure is achieved using a multiscale MLP, which allows the network to handle different scales of detail simultaneously. As a result, Mip-NeRF achieves better accuracy than NeRF, especially in scenes viewed at various resolutions. It reduces the error rate by 60\% compared to NeRF and is faster by 7\% with only half the number of parameters. This is achieved by using a scale-aware structure and merging the separate ``coarse'' and ``fine'' MLPs of NeRF into a single MLP. This innovation has the added benefit of reducing the complexity of the model and the computational overhead, making Mip-NeRF more efficient than its predecessor. Figure \ref{fig:mipnerf} shows the comparison of NeRF and Mip-NeRF neural rendering methods. 

Point-NeRF \cite{pointNerf} distinguishes itself from Mip-NeRF by addressing the computational challenges of training and rendering NeRFs through a point-based approach. Instead of representing scenes as continuous volumes, Point-NeRF represents them as collections of points, each associated with a neural network that predicts its color and appearance. This point-based representation is more efficient than traditional NeRFs, as NeRFs use continuous volume representations. 
The authors propose a pipeline for efficiently reconstructing point-based radiance fields using a neural network. The pipeline involves generating and optimizing a point-based field per scene, employing point growing and pruning techniques. Point pruning eliminates unnecessary outlier points based on confidence values, while point growing adds new points to cover missing scene geometry in empty regions. A feed-forward neural network facilitates efficient reconstruction, and a neural generation module predicts all neural point properties, including point locations, features, and confidence. Deep MVS \cite{furukawa2015multi} methods, utilizing cost volume-based \cite{Introduc11:online} 3D CNNs, are used to generate 3D point locations, while a 2D CNN extracts neural 2D image feature maps from each image. The final neural point cloud is obtained by combining point clouds from multiple viewpoints, and the point generation networks and representation networks are trained end-to-end with a rendering loss. This pipeline significantly reduces per-scene fitting time and achieves high-quality rendering.
The experiments demonstrate that Point-NeRF achieves state-of-the-art results on multiple datasets and effectively handles errors and outliers through the proposed pruning and growing mechanism.

Training NeRF involves a slow per-scene optimization process, making it time-consuming, particularly for large-scale scenes. Additionally, as a neural network, NeRF has limited capacity to represent intricate 3D geometry and appearance, posing challenges for capturing fine details in large-scale scenes. NeRFusion \cite{NeRFusion}  addresses these limitations by proposing a new method for reconstructing large-scale scenes. NeRFusion fuses radiance fields from a sequence of images to create a global scene representation. It predicts per-frame local radiance fields using a neural network and fuses them using a recurrent neural network. The resulting representation can be fine-tuned for better renderings. A rendering pipeline is introduced, trained with ground truth images, without additional geometry supervision. Pre-training is performed on the local reconstruction network and radiance field decoder. The entire pipeline is trained jointly, yielding high-quality radiance fields and realistic renderings. A global neural volume fusion network with gated recurrent units (GRU) and sparse 3D CNNs is used to fuse local feature volumes into a global volume. The framework is generalizable and efficient, utilizing MLP networks for regression and modeling volumes in canonical world space. Voxel pruning optimizes memory and rendering efficiency.

Traditional NeRFs also lack the ability to deform as they are trained on a fixed set of images and are incapable of updating to reflect changes in the scene. In contrast, DRF-Cages \cite{Deforming} allows for deformation by manipulating a triangular mesh known as the ``cage''. 
The process involves constructing the cage around the object of interest and deforming it using a mesh deformation algorithm. The radiance field is then updated to reflect the new shape of the object. Cage generation involves converting the optimized radiance field into a fine mesh and creating the corresponding cage. However, complex shapes or fine details may require manual refinement of the automatically generated cages.
Points inside the cage are represented by cage coordinates, which are calculated based on their relative positions to the cage vertices. By manipulating the cage vertices, the cage can be deformed, and the radiance field is deformed accordingly. A deformed-to-canonical mapping is used to compute color and density values from the canonical radiance field for points inside the deformed cage.
While the proposed method shows promising results, it faces challenges such as representing detailed cage shapes with a small number of vertices and effectively generating cages from 3D scenes. Failure cases include artifacts caused by occlusion or drastic deformations.

Table \ref{tab:1} offers a comparative view of different approaches to improve the rendering quality of NeRF's algorithm. This table facilitates an evaluation of the trade-offs between these approaches with respect to their strengths and weaknesses.

\begin{table}[htbp]
\caption{\label{tab:1}Comparison of approaches in terms of improvement of rendering quality}
\centering
\resizebox{\textwidth}{!}{
\begin{tabular}{|l|l|l|l|}
\hline
Paper & Key Features & Strengths & Weaknesses \\ \hline
Mip-NeRF \cite{mipNerf} &
  \begin{tabular}[c]{@{}l@{}}Uses a multiscale representation to reduce \\ aliasing artifacts and improve the ability \\ to represent fine details\end{tabular} &
  \begin{tabular}[c]{@{}l@{}}More efficient and \\ accurate than NeRF\end{tabular} &
  \begin{tabular}[c]{@{}l@{}}More complex and requires \\ more training data\end{tabular} \\ \hline
Point-NeRF \cite{pointNerf} &
  \begin{tabular}[c]{@{}l@{}}Uses a point cloud to represent the scene \\ geometry, which can be pre-trained on \\ a dataset of 3D scenes\end{tabular} &
  \begin{tabular}[c]{@{}l@{}}Robust against errors and\\outliers and faster than NeRF\end{tabular} &
  \begin{tabular}[c]{@{}l@{}}Less accurate than NeRF \\ for scenes with complex geometry\end{tabular} \\ \hline
NeRFusion \cite{NeRFusion} &
  \begin{tabular}[c]{@{}l@{}}Fuses multiple NeRFs together to achieve \\ higher resolution and accuracy\end{tabular} &
  \begin{tabular}[c]{@{}l@{}}Capable of handling occlusions \\ and truncations\end{tabular} &
  \begin{tabular}[c]{@{}l@{}}Difficulty with scenes having distant\\ backgrounds and foreground objects\end{tabular} \\ \hline
DRF-Cages \cite{Deforming} &
  \begin{tabular}[c]{@{}l@{}}Uses a cage to deform a NeRF, which can \\ be used to create new shapes and textures\end{tabular} &
  \begin{tabular}[c]{@{}l@{}}Capable of creating more \\ complex and realistic scenes\end{tabular} &
  \begin{tabular}[c]{@{}l@{}}Difficulty in representing detailed cage\\shapes with a small number of cage vertices\end{tabular} \\ \hline
\end{tabular}}
\end{table}

\section{Scalability}
NeRF training involves optimizing a complex neural network to learn the scene's volumetric representation and view-dependent radiance for each ray. It is computationally intensive and time-consuming, particularly with high-resolution or intricate scenes, taking hours to days depending on the dataset and resources. Due to these challenges, NeRF is not well-suited for real-time or interactive applications. Researchers have been working on improving its efficiency, exploring variants and extensions to make it more practical for real-world scenarios. Table \ref{tab:2} comprehensively compares various approaches in terms of their scalability, particularly in managing computational demands for different scenarios. 

\begin{table}[hbtp]
\caption{\label{tab:2}Comparison of approaches in terms of scalability}
\centering
\resizebox{\textwidth}{!}{\begin{tabular}{|l|l|l|l|}
\hline
Paper &
  Key Features &
  Strengths &
  Weaknesses \\ \hline
FastNeRF \cite{FastNeRF} &
  \begin{tabular}[c]{@{}l@{}}Uses a novel sampling strategy \\ to achieve high frame rates\end{tabular} &
  \begin{tabular}[c]{@{}l@{}}Produces high-fidelity \\ images at 200fps\end{tabular} &
  \begin{tabular}[c]{@{}l@{}}Requires more training \\ data than other methods\end{tabular} \\ \hline
KiloNeRF \cite{KiloNeRF}&

  \begin{tabular}[c]{@{}l@{}}Uses a hierarchical representation \\ of the scene to reduce the number \\ of parameters\end{tabular} &
  \begin{tabular}[c]{@{}l@{}}Very efficient, can  train \\on a single GPU in \\a few hours\end{tabular} &
  \begin{tabular}[c]{@{}l@{}}Produces lower-quality \\ images than other methods\end{tabular} \\ \hline
Block-NeRF \cite{blocknerf}&
  \begin{tabular}[c]{@{}l@{}}Divides the scene into blocks and \\ renders each block independently\end{tabular} &
  \begin{tabular}[c]{@{}l@{}}Scales to very large \\ scenes\end{tabular} &
  \begin{tabular}[c]{@{}l@{}}Requires more memory \\ than other methods\end{tabular} \\ \hline
Mega-NeRF \cite{MegaNeRF}&
  \begin{tabular}[c]{@{}l@{}}Uses a dynamic grid that is adapted to\\ the scene being rendered\end{tabular} &
  \begin{tabular}[c]{@{}l@{}}Produces high-quality \\ images of large scenes\end{tabular} &
  \begin{tabular}[c]{@{}l@{}}Very computationally \\ expensive\end{tabular} \\ \hline
MobileNeRF \cite{MobileNeRF}&
  \begin{tabular}[c]{@{}l@{}}Exploits the polygon rasterization pipeline \\ to render NeRFs on mobile devices\end{tabular} &
  \begin{tabular}[c]{@{}l@{}}Very fast on mobile \\ devices\end{tabular} &
  \begin{tabular}[c]{@{}l@{}}Produces lower-quality \\ images than other methods\end{tabular} \\ \hline
\end{tabular}}
\end{table}

\begin{figure}[htbp]
\begin{center}    \includegraphics[width=0.98\textwidth]{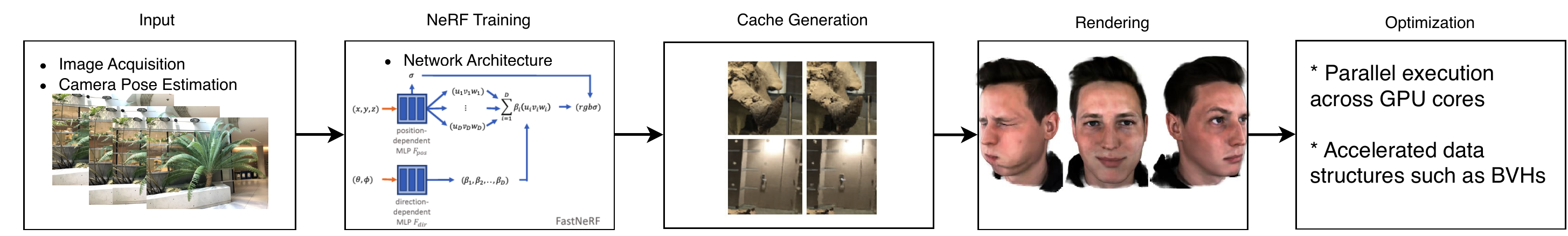}
\caption{The pipeline consists of five key stages. First, a multi-camera rig is used to capture images of a scene from varied viewpoints. The camera pose is estimated for each image view. These images and poses are then utilized to train a FastNeRF model that represents the scene as an implicit neural representation. Next, the positions and directions are densely sampled from the trained model, and the radiance field outputs are cached in a sparse 3D grid structure. At render time, rays are traced from the camera through pixels, and the radiance field is queried from the cache at intersection points. Finally, optimizations such as parallel GPU execution and accelerated data structures enhance performance.}
\label{fig:fastnerf}
\end{center}
\end{figure}

FastNeRF \cite{FastNeRF} is an efficient and easy-to-train variant of NeRF that achieves high-quality rendering at an unprecedented 200 frames per second (FPS). FastNeRF introduces a novel architecture that divides the original NeRF neural network into two separate networks, each dependent on positions and directions, respectively. The output of these two functions is efficiently cached, leading to a remarkable three orders of magnitude improvement in rendering performance without compromising visual quality. Remarkably, while the standard NeRF model demands an impractical 5,600 Terabytes cache size, FastNeRF dramatically reduces this requirement to a manageable 54 GB, making it suitable for consumer-grade hardware. Furthermore, the memory requirement can be further lowered by using smaller cache sizes, rendering FastNeRF even more feasible for resource-constrained environments. The choice of $k$ and $l$ (the numbers of samples for position and direction) depends on scene size and image resolution. FastNeRF offers a versatile caching mechanism, significantly outperforming the original NeRF by being 3,000 times faster and at least an order of magnitude faster than existing alternatives while preserving visual quality. FastNeRF's factorization of the original NeRF into two separate functions offers additional advantages, such as reducing the numbers of network parameters, improving network control, and simplifying the training process. Moreover, FastNeRF can easily accommodate other techniques, including multi-resolution representations and volume rendering, making it a versatile tool for various applications. Figure \ref{fig:fastnerf} illustrates the end-to-end pipeline for FastNeRF. 

In the pursuit of real-time rendering for Neural Radiance Fields (NeRF), recent studies have explored converting the neural representation into a discrete one after training \cite{FastNeRF, abs-2103-14645, yu2021plenoctrees}. However, these methods often suffer from significant GPU memory overhead, limiting their applicability to larger scenes. To address these challenges, KiloNeRF  \cite{KiloNeRF} presents a novel approach that harnesses thousands of small Multi-Layer Perceptrons (MLPs) to accelerate the rendering process while synthesizing high-quality novel views of a scene. KiloNeRF builds upon NeRF by representing the scene as a collection of volumetric density and color values using MLPs. Each scene is subdivided into a uniform resolution grid, with independent MLPs assigned to specific 3D cells, thereby optimizing GPU memory usage and computation time. Knowledge distillation is employed to train the KiloNeRF model, starting from an ordinary NeRF model and fine-tuning it with a photometric loss on training images. To maintain visual quality and avoid artifacts in empty spaces, KiloNeRF applies regularization strategies to the weights and biases of the last two layers of the network responsible for view-dependent color modeling. Additionally, the method employs equidistant point sampling along the rays to further reduce the number of points evaluated by the network. This process is complemented by Empty Space Skipping (ESS) and Early Ray Termination (ERT) techniques. ESS utilizes an occupancy grid with binary values to indicate whether a cell contains content, ensuring that the network is only evaluated when the cell is occupied. ERT terminates ray evaluation when the transmittance value approaches zero, conserving computation time. The prototype implementation of KiloNeRF is based on PyTorch \cite{NEURIPS2019_bdbca288}, MAGMA \cite{abdelfattah2017novel}, and Thrust, incorporating custom CUDA kernels for various tasks such as sampling, empty space skipping, early ray termination, positional encoding, network evaluation, and alpha blending. A custom routine is developed to fuse the entire network evaluation into a single CUDA kernel, efficiently handling simultaneous queries of thousands of networks. To optimize performance, MAGMA is employed to manage multiple network queries effectively. While the proposed KiloNeRF method has demonstrated impressive results, enabling the rendering of high-quality novel views of a scene faster than previous approaches, it should be noted that the method assumes a bounded scene. Addressing this limitation and extending the method to handle unbounded scenes could be a promising direction for future research in this field.

In the realm of large-scale 3D scene synthesis, the recent paper Block-NeRF \cite{blocknerf} introduces a groundbreaking approach that overcomes the memory limitations of traditional NeRF. By dividing the scene into manageable blocks, Block-NeRF presents a novel NeRF variation capable of representing expansive environments. To tackle the challenges posed by rendering large-scale scenes, Block-NeRF adopts a unique strategy. The scene is divided into individually trained Block-NeRFs, which are dynamically rendered and combined during inference. The critical aspect here is the placement and size of these blocks to achieve comprehensive coverage of the target environment. The authors propose an effective heuristic involving the placement of blocks at intersections, where each block covers 75\% of the distance to the next intersection. This configuration ensures a 50\% overlap between adjacent blocks, facilitating appearance alignment. Additional blocks can be introduced as needed to connect intersections, and a geographical filter is employed to maintain appropriate training data within each block's intended bounds. This design affords maximum flexibility, enables scalability to arbitrarily large environments, and facilitates piecewise updates or introductions of new regions. During view synthesis, Block-NeRF selectively renders a subset of Block-NeRFs based on their geographical location relative to the camera. To ensure visual consistency across different Block-NeRFs, appearance matching is performed, and the optimized appearance is iteratively propagated through the scene. The process starts from the root Block-NeRF, with consideration of multiple surrounding blocks when computing the loss. Notably, the authors discover that incorporating the camera's exposure information into the appearance prediction section effectively mitigates visual discrepancies, enhancing the quality of the synthesized views. However, some limitations are identified in the Block-NeRF model. During training, the model filters out transient objects via masking, resulting in potential artifacts in the rendered outputs. Additionally, challenges persist in sampling distant objects with lower density, leading to blurrier reconstructions.

While BlockNeRF uses a fixed grid of blocks,   Mega-NeRF \cite{MegaNeRF} uses a dynamic grid that is adapted to the scene being rendered. This makes Mega-NeRF a more scalable and efficient framework for large-scale visual captures using NeRF. The proposed framework addresses several challenges related to scaling up NeRFs for large-scale scenes, including handling thousands of images with varying lighting conditions, accommodating large model capacities, and mitigating slow rendering speeds. Mega-NeRF adopts a sparse network structure and a geometric clustering algorithm to train and render large-scale scenes. The scene is decomposed into cells with centroids, and a two-stage hierarchical sampling procedure with positional encoding captures high-frequency details. By generating opacity and color using model weights nearest to the query point, Mega-NeRF efficiently generates realistic representations of the scene. To optimize training efficiency, each Mega-NeRF submodule can be trained independently in parallel, without requiring inter-module communication, thanks to their self-contained MLP nature. Moreover, only possibly relevant pixels that cross spatial cells are added to the trainset for each submodule, minimizing the trainset size. A small overlap factor between cells helps reduce visual artifacts near boundaries. Additionally, pruning irrelevant pixels/rays contributes to further reduction of trainset size. For interactive rendering, Mega-NeRF introduces a novel method that involves caching and temporal coherence to speed up the process. Precomputing a coarse cache of opacity and color and dynamically subdividing the tree during interactive visualization enable quick initial view production and subsequent model sampling for image refinement. The final round of guided ray sampling further enhances rendering quality. The process is accelerated by rendering rays in a single pass and accumulating transmittance along the ray. Extensive evaluation of existing datasets and drone footage shows notable improvements in training speed and Peak Signal-to-Noise Ratio (PSNR) compared to other NeRF approaches. Introducing a novel dataset comprising high-definition images collected through drone recordings, covering a vast area of 100,000 square meters around an industrial complex and containing thousands of images, demonstrates Mega-NeRF's capability to handle large-scale visual captures.

The MobileNeRF \cite{MobileNeRF} introduces a novel encoding scheme that addresses the limitations of traditional NeRF rendering methods, presenting a more efficient approach for rendering neural fields on mobile architectures with constrained memory resources. The key innovation lies in the utilization of textured polygons, which significantly reduces the data requirements while preserving rendering quality. By leveraging the polygon rasterization pipeline, MobileNeRF achieves a more compact scene representation, enabling it to render complex scenes on mobile devices efficiently. The representation consists of textured polygons with associated opacity and feature vectors stored in a texture atlas. The rendering process is executed by a lightweight MLP running in a GLSL (OpenGL Shading Language) fragment shader \cite{abs-2103-14645}, which yields the final output color. A fixed mesh topology and three optimized MLPs are employed to train the system to minimize the mean squared error between predicted and ground truth colors of training image pixels. The predicted color is computed through alpha compositing using radiance and opacity, both obtained by evaluating the MLPs at specific positions. Subsequently, the representation is converted to a polygonal mesh after binarization and fine-tuning, allowing only partially visible quads to be stored. Creating the texture map involves generating a texture image with each quad allocated a patch. Discrete opacity and features are then baked into the texture map through pixel-wise iteration, converting pixel coordinates to the corresponding 3D coordinates. The proposed rendering system is highly optimized for GPUs, ensuring the efficient execution of the entire pipeline on various mobile devices, even achieving interactive frame rates.

\section{High-Quality Large Dataset}
NeRF requires a large dataset of high-quality images, which can be challenging to obtain and affect the quality of synthesized views. Researchers have tried to address this issue by adding prior conditions to the model, which can assist in image reconstruction or be used in a generative fashion to produce new images. Additionally, one-few-zero shot learning can generate 3D models with limited or no data, making NeRF useful in real-world applications. Priors can improve neural view synthesis and enable the reconstruction of images from sparse collections. In a generative manner, priors can generate new images based on prior information.

\subsection{Prior-based Approaches}
CAMPARI \cite{CAMPARI} stands out as a versatile approach for photorealistic image synthesis using deep generative models. Unlike other 3D-aware methods that necessitate camera modeling, CAMPARI offers a more principled solution by simultaneously learning a camera generator and an image generator. This approach enables the use of complex camera distributions without the need for manual tuning. The scene is efficiently represented through foreground and background decomposition, leading to 3D-consistent representations and accurate camera distribution recovery.
A key advantage of CAMPARI is its ability to generate new scenes with precise control over camera viewpoint, shape, and appearance during testing. To achieve this, stratified sampling and prior knowledge injection approximate numerical integration, enhancing the quality of the generated images.
Owing to its remarkable performance without camera parameter tuning, CAMPARI surpasses baseline models, which heavily depend on accurately matched camera distributions.
However, CAMPARI faces the challenge of generating inward-facing faces compared to the traditional ``inverted faces'' also known as the ``hollow face illusion'' where convex faces appear concave. To address this issue, the authors are actively exploring the incorporation of stronger 3D shape biases into the generator model, aiming for even more reliable results.

While CAMPARI is better at generating more photorealistic 3D representations, CoCo-INR \cite{cocoinr} is better at generalizing to novel views. Existing methods require a dense set of calibrated views to produce high-quality 3D scenes, which makes them less effective with fewer input views. To overcome this limitation, CoCo-INR proposes injecting prior information into the coordinate-based network to enhance the feature representation and reduce the dependence on massive, calibrated images. The proposed approach employs two attention modules: codebook attention and coordinate attention. The codebook attention extracts valuable prototypes from the codebook, while the coordinate attention enables each coordinate to query representative features from the prototypes. Integrating prior information through these attention mechanisms results in more accurate and efficient 3D representations. The CoCo-INR approach has been applied to multi-view scene reconstruction and novel view synthesis. The authors have used a combination of geometry and appearance networks to predict the scene, using the NeRF++ \cite{DBLP:journals/corr/abs-2010-07492} framework. The networks are trained by minimizing the difference between the rendered colors and the ground truth colors without 3D supervision. The injection of prior information through CoCo-INR provides several advantages over existing methods, such as reducing the dependency on calibrated views, enhancing the quality of 3D representation, improving the accuracy of the predicted scenes, and using attention mechanisms to provide a more efficient way to integrate prior information.

CoCo-INR requires a codebook prior, which makes the model still complex and difficult to train. DietNerF \cite{Diet} further reduces the complexity and computation cost by incorporating prior knowledge from a pre-trained image encoder to guide the optimization process of NeRF. This prior knowledge comes in the form of a semantic consistency loss, which ensures that the high-level semantic features of the observed and the rendered views are similar. Semantic consistency loss allows the DietNeRF model to capture stable high-level semantics across different viewpoints. The authors evaluate two sources of supervision for representation learning in DietNeRF: a pre-trained visual encoder such as the CLIP (Contrastive Language-Image Pre-Training) \cite{radford2021learning} model and visual classifiers pre-trained on ImageNet images. The CLIP model produces normalized image embeddings, while the visual classifiers use Vision Transformer architectures that extract features from image patches and produce a single, global embedding vector. The CLIP model used in DietNeRF is trained on millions of images with captions, providing rich supervision for image representations. The captions provide semantically sparse and dense learning signals, which helps the image representation capture fine-grained details and high-level semantics. During training, the DietNeRF model is improved for efficiency and quality by using low-resolution semantic consistency renderings and sampling poses from a continuous distribution, Local Surface Consistency (LSC) \cite{375159:online} minimization, and mixed precision computation and memory-saving techniques. These optimizations allow the model to produce high-quality 3D reconstructions from a small number of input images while also reducing the computational requirements of the model.

Both DietNeRF and DreamFields \cite{DreamFields} are proposed by the same authors. However, DietNeRF can only generate 3D objects from images, while DreamFields can generate 3D objects from natural language descriptions. Natural language descriptions can help reduce the data dependencies by providing more information about the desired output, regularizing the model, and preventing overfitting. This makes Dream Fields more versatile and applicable to a wider range of tasks. The method employs image-text models trained on large, captioned image datasets to guide the generation process. The method optimizes a NeRF for high scores with a target caption based on a pre-trained CLIP model \cite{radford2021learning}. Dream Fields trains a radiance field that renders images with high semantic similarity to a given text prompt, using pre-trained image-text retrieval models. The approach also incorporates simple geometric priors to improve fidelity and visual quality. An MLP optimizes the approach to produce outputs representing a scene's differential volume density and color at every 3D point, solely dependent on the 3D coordinates instead of the camera's viewing direction. The method uses segments spaced at equal intervals with random jittering along the ray to compute the transmittance for rendering an image. To train Dream Field in a zero-shot setting, the approach randomly samples pose and uses a CLIP network to measure the similarity between the rendered image and the provided caption. The image and text encoders used in the CLIP network are from CLIP and a baseline Locked Image-Text Tuning (LiT) ViT B/32 model \cite{zhai2022lit}, both trained contrastively on large datasets of captioned images. The proposed method supports 3D data augmentations by uniformly sampling camera azimuth in 360 degrees around the scene, and camera elevation, focal length, and distance from the subject can also be augmented. To improve coherence and reduce artifacts and spurious density, the method regularizes the opacity of Dream Field renderings by maximizing the average transmittance of rays passing through the volume. The transmittance loss is defined by the probability of light passing through $N$ discrete segments of the ray and is compared to baseline sparsity regularizers. The Dream Fields approach can produce realistic, multi-view, consistent object geometry and color from various natural language captions despite diverse, captioned 3D data scarcity. However, iterative optimization can be expensive, and using the same prompt for all perspectives can result in repeated patterns on multiple sides of an object. The image-text models used to score renderings are imperfect and can inherit harmful biases. Dream Fields does not target complex scene generation or handle scene layout. 

\subsection{GAN-based Approaches}

Efforts have been made to enhance NeRF's ability to generalize to new scenes with limited views, but these approaches still rely on multiple views. Addressing this challenge, Pix2NeRF \cite{Pix2NeRF} introduces a novel pipeline to generate NeRF from a single input image. It builds upon $\pi$-GAN \cite{DBLP:journals/corr/abs-2012-00926}, a 3D image synthesis generative model that maps latent codes to radiance fields. Pix2NeRF employs an unsupervised approach and can be trained with independent images, offering potential applications in 3D avatar generation, novel view synthesis, and super-resolution. However, a limitation of Pix2NeRF is its current restriction to work with only one category per dataset, impeding its ability to generalize to new categories. For example, if the Pix2NeRF method is trained on a dataset of images of cars, it will only be able to generate NeRF representations of cars but not of any other object. To overcome this limitation, alternative research directions are proposed, including local conditional fields similar to PixelNeRF \cite{14}, enabling generalization to unseen categories, multi-instance, and real-world scenes. 
Pix2NeRF is built on top of $\pi$-GAN but is not limited to using $\pi$-GAN as its backbone. Newer generative NeRF models such as EG3D \cite{DBLP:journals/corr/abs-2112-07945} may achieve better visual quality if used as a backbone. Despite the challenging nature of architecture search, especially concerning the encoder, the authors suggest adopting more mature encoder architectures from 2D GAN feed-forward inversion literature, such as pixel2style2pixel \cite{abs-2008-00951}. This adaptation could significantly improve Pix2NeRF's performance. The authors anticipate that Pix2NeRF will be a strong baseline for future research in this domain, having demonstrated its superiority over naive GAN inversion methods through extensive ablation studies. 

VoxGRAF \cite{VoxGRAF} enhances the Pix2NeRF approach by utilizing sparse voxel grid representations for efficient and consistent generative modeling for the 3D scene instead of an MLP. This results in faster rendering speeds and improved 3D consistency compared to methods like Pix2NeRF, that rely on neural rendering. The authors replace the coordinate-based MLP in the 3D generator with a sparse 3D Convolutional Neural Network (CNN). The model consists of a 3D foreground generator and a 2D background generator. The foreground generator is based on the StyleGAN2 \cite{karras2020analyzing} architecture and uses a latent code and camera pose to map color and density values onto a sparse voxel grid. The background generator maps the latent code to a background image. Finally, the two images are combined using alpha composition to create the final image. Alpha composition is a technique for compositing multiple images together, where each image is assigned with an alpha value that determines its transparency. The authors compared their method with the state-of-the-art baselines and found that their approach resulted in more consistent multi-view results. VoxGRAF has the potential to be faster and more efficient, leading to more accurate and consistent 3D generative models, especially in real-time applications. 

VoxGRAF is faster than HyperNeRFGAN \cite{HyperNeRFGAN}  because it does not need to create a detailed NeRF representation of the scene. However, HyperNeRFGAN can create more realistic images because it has a more detailed scene representation compared to VoxGRAF. The authors present a new approach to 3D object representation that addresses the limitations of standard voxel and point cloud methods. The authors propose using NeRFs to synthesize 3D scenes from a small set of 2D images. Their proposed  HyperNeRFGAN model, combines the hypernetworks paradigm and NeRF representation. The generator takes a sample from a base distribution and returns the parameters for the NeRF model, which transforms spatial locations into emitted colors and volume density. The model is trained using the StyleGanv2 \cite{karras2020analyzing} objective loss function, and the noise vector is transformed to obtain the weights of the NeRF model. The generator produces 2D images with a 3D-aware NeRF representation to ensure accurate 3D object creation. However, one limitation is that it uses only 2D images and not 3D information. Future work includes adding 3D mesh structure information to improve the model's capabilities.

\begin{figure}[htbp]
\centering
\includegraphics[width=\linewidth]{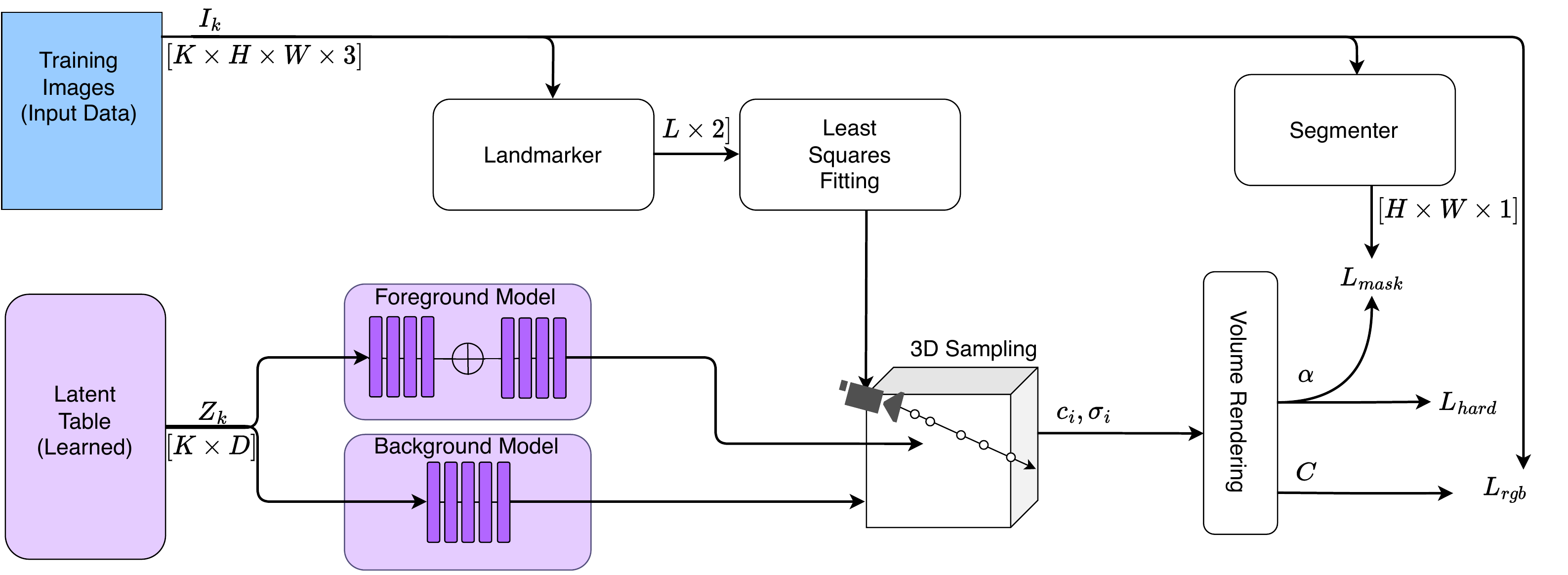}
\caption{LOLNeRF simultaneously learns a table of latent codes for each image and foreground and background neural radiance fields (NeRFs). The volumetric rendering output is evaluated against each training pixel using a per-ray RGB loss and against an image segmenter using an alpha value loss. Camera alignments are determined by fitting the 2D landmark outputs to class-specific canonical 3D keypoints using a least-squares approach.}
\label{fig:lolnerf}
\end{figure}

LOLNeRF \cite{Lolnerf} is a more straightforward approach than HyperNeRFGAN. Because it only uses a single image to learn a NeRF representation of the scene. However, the simplicity of LOLNeRF is also one of its strengths. LOLNeRF is much faster than HyperNeRFGAN, which makes it a good choice for applications where speed is important. This method allows the 3D structure to be rendered from different views without requiring multi-view data. To achieve this, all images in the dataset are aligned to a canonical pose, and a shared generative model with an autoencoder framework \cite{deepsdf} is used. This allows the method to be trained with arbitrary image sizes by separating training complexity from image resolution. This method outperforms adversarial methods in representing object appearance and predicting geometry without geometric supervision. To improve generalization, two models are trained, one for foreground objects and one for background. The method models shapes as solid surfaces, enhancing the quality of predicted shapes. The training process is optimized to reconstruct images from datasets and find optimal latent representations for each image without rendering entire images or patches. Camera parameters are estimated using a pre-trained network that extracts 2D landmarks from the images. The model is capable of generating new 3D reconstructions given a pre-trained model and a latent code. It can be trained with arbitrary image sizes without increasing memory requirements during training and outperforms adversarial methods in representing the appearance of objects from the learned category. However, LOLNeRF relies on other methods for semantic information extraction and can lead to failure cases when the estimated pose or segmentation needs to be corrected. While the auto-decoder framework offers benefits compared to GANs, it lacks the ability to optimize the realism of images. Future work aims to improve image quality by adding adversarial training to the existing method. The method can potentially augment image quality and improve the perceptual quality of images rendered from novel latent codes. This could be a possible direction for future work. Figure \ref{fig:lolnerf} shows the architecture of LOLNeRF. 

\section{Representing Articulated Objects}
Dealing with articulated objects in NeRF is a challenging problem in computer vision research. While NeRF has shown promising results in representing static scenes, it struggles to capture the dynamic and complex nature of articulated objects such as humans or animals. One of the main reasons for this is NeRF's assumption of a static scene represented as a continuous volumetric function, which fails to consider the motion and deformation of the object's geometry. Additionally, modeling each part of an articulated object explicitly using a set of parametric functions is impractical and hence requires new approaches for representation. Articulated object recognition is an active area of research, and recent works are applying the concept of NeRF for articulated models of people or cats, representing the shape of the object in the image using interconnected rigid parts.

iNeRF \cite{iNeRF} introduces a new framework for 6 DoF (degrees of freedom) pose estimation, which uses an observed image, an initial pose estimate, and a 3D object or scene as input to estimate the camera pose. Unlike traditional NeRF models that optimize weights using a set of given camera poses and image observations, iNeRF aims to recover the camera pose given the weights and the image by inverting the trained NeRF. The NeRF model's ability to render an image observation by taking an estimated camera pose is utilized to minimize the photometric loss function and update the pose. The authors found that sampling all pixels in the image is not feasible due to memory limitations and explored three strategies for selecting a smaller set of rays to compute the loss function. Random sampling is found to be ineffective when the batch size of rays is small, as most randomly sampled pixels correspond to textureless regions. The authors propose interest region sampling, which samples from dilated masks centered on the interest points and find that it speeds up the optimization when the batch size of rays is small. However, iNeRF has limitations as it does not model lighting or occlusion, negatively affecting its performance. One possible solution proposed in the paper is to include appearance variation with transient latent codes and optimize these codes along with the camera pose in iNeRF. This method takes approximately 20 seconds to run 100 optimization steps, making it unsuitable for real-time use. Improvements in NeRF's rendering speed could mitigate this issue.

The NeRF model faces challenges in representing articulated objects due to the complex relationship between kinematic \cite{Kinemati95} representation and the radiance field. While iNeRF offers a detailed volumetric representation, it comes with a high computational cost, whereas Neural Articulated Radiance Field's (NARF) \cite{NARF} part-based approach sacrifices some details but gains efficiency and better generalization capabilities. NARF \cite{NARF} uses a neural network to transform the 3D location and 2D viewing direction into density and a RGB color value. The density controls the radiance of a ray passing through a location, and the network consists of two ReLU MLP networks, one for volume density and another for RGB colors. A rigidly transformed neural radiance field (RT-NeRF) is used to model each part, and a single unified NeRF is trained to encode multiple parts while avoiding the part dependency issue. The authors present two basic solutions for NARF: Part-Wise NARF ($NARF_P$) and Holistic NARF ($NARF_H$), and analyzed their pros and cons. Part-Wise NARF decomposes the object into parts and trains a separate RT-NeRF for each part. This approach allows for better handling of part dependencies but can be computationally expensive. Holistic NARF models the entire object as a single RT-NeRF, which is computationally efficient but can lead to issues with implicit transformations. The authors propose a final solution, called Disentangled NARF, which combines the advantages of both Part-Wise and Holistic NARF. Disentangled NARF decomposes the object into parts, but the RT-NeRF for each part is trained in a common latent space, allowing for better handling of part dependencies while avoiding implicit transformations. The network uses a hierarchical volume sampling strategy to estimate the densities and colors of samples and then renders the color and mask of each camera ray using volume rendering.

\begin{figure}[htbp]
\centering
\includegraphics[width=\linewidth]{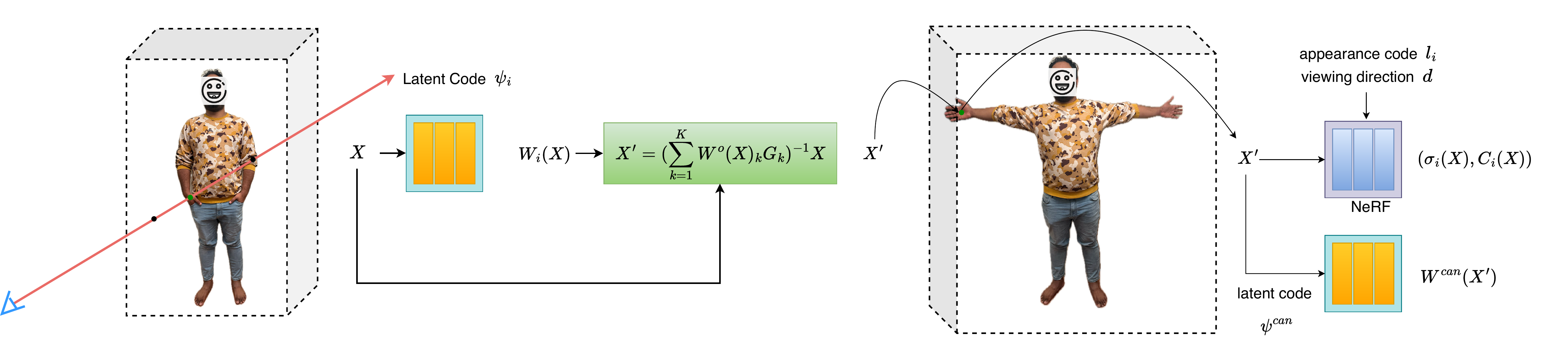}
\caption{Given a query point $X$ in observation space at frame $i$, Animatable NeRF infer its blend weight $W_i(X)$ using a neural blend weight field conditioned on $\psi_i$. Based on $W_i(X)$ and the skeleton, Animatable NeRF obtains the corresponding canonical point $X^\prime$ using the equation shown in the figure. Given $X^\prime$, view $d$, and appearance code $l_i$, the template NeRF model predicts density and color. to animate the template $W_{can}(X^\prime)$ is obtained at the canonical space .}
\label{fig:animatable}
\end{figure}

Animatable NeRF \cite{Animatable} is a recent advancement of NeRF that extends its capabilities to handle dynamic scenes. The method leverages the strengths of NeRF and the parameterized human model Skinned Multi-person Linear Model (SMPL) \cite{loper2015smpl} to create high-quality human reconstructions. The authors optimize NeRF and SMPL parameters for better results and faster convergence by introducing pose-guided deformation and analysis-by-synthesis.
The method maps 3D position, shape, and pose into color and density. The authors use the SMPL model to model human appearance and geometry and the NeRF model to represent the spatial distribution of light in the scene. To handle human movements between different frames, the authors transform the 3D position in the observation space into canonical space using the SMPL model. This allows the method to generalize to unseen poses without requiring a large amount of training data.
The authors use volume rendering techniques to render the NeRF into a 2D image and obtain pixel colors by accumulating the colors and densities along the corresponding camera ray. They approximate the continuous integration by sampling points between the near and far planes along the camera ray.
The proposed method learns an animatable NeRF for human subjects by explicitly deforming the observation space under the guidance of SMPL transformations. The authors address the problem of inaccurate human body estimation by fine-tuning the SMPL parameters during training using VIBE (Video Inference for Human Body Pose and Shape Estimation) \cite{kocabas2020vibe}.
Animatable NeRF is a promising new approach for modeling and animating dynamic human bodies. The method has several advantages over previous methods, including its ability to handle complex poses, scalability, and potential for real-time applications. Figure \ref{fig:animatable} shows an overview of the Animatble NeRF method that reconstructs an animatable human model from a multi-view video.

HumanNeRF \cite{HumanNeRF} is a promising approach for reconstructing animatable human models from monocular videos. It is a simpler representation than Animatable NeRF and can be used with a single view. This is achieved through a modified version of NeRF that incorporates human motion and can be generalized.
The approach first blends multi-view input images using an aggregated pixel alignment feature. This results in a model that can predict the volume density and color at a given point before deformation using Dynamic Human Volume Rendering \cite{10.1145/800031.808594}.
A non-rigid deformation field is then used to learn subtle displacement, which warps the human body from the current time frame to a common canonical pose. This allows the model to be generalized to unseen people.
The authors suggest a fast fine-tuning solution to address the issue of limited training data and diversity among different identities and scenes. The network is trained on various human subjects/performers, and the feature blending network is frozen. When given an unseen subject, the network parameters of the deformation field and the generalizable NeRF are optimized.
A neural blending scheme is also introduced to refine the textures produced by NeRF rendering in multi-view settings. This is done by first rendering a depth map from the target view, and each point in the map is back-projected into the neighboring views to fetch the colors and visibility.
The generalizable NeRF module and the appearance blending network are trained separately. The network is optimized using a color loss and a silhouette loss, with the total loss combined.
The HumanNeRF approach is promising for generating high-quality and photo-realistic images of dynamic humans using sparse RGB streams. 

At around the same time, BANMo \cite{BANMo} introduces a novel approach to generating high-quality, animatable 3D models of flexible objects, focusing on realistic images of humans. The method employs implicit neural functions to represent 3D geometry and appearance in a canonical space and a neural blend skinning model to constrain object deformation. By using Multilayer Perceptron (MLP) networks, the method predicts properties of 3D points in the canonical space, enabling the matching of pixels across various viewpoints and lighting conditions. BANMo incorporates time-dependent warping functions, volume rendering, and 2D flow projection to account for object deformation. Articulated body motion is approximated by blending rigid transformations of 3D bone coordinates, with poses represented using angle-axis rotations and 3D translations. Skinning weights are assigned to bones based on a conditioning skinning weight function. The approach leverages canonical feature embeddings to jointly optimize shape, articulation, and embeddings. The canonical embeddings are self-supervised to ensure consistency between feature matching and geometric warping, facilitating robust reconstruction across multiple videos. Optimization involves minimizing reconstruction losses, geometric feature registration losses, and a 3D cycle consistency regularization loss. While the method shows promising results, there is scope for future work to improve computational efficiency during optimization.

Table \ref{tab:3} presents a comparative analysis of various approaches in the context of articulation, focusing on their contributions to free-viewpoint rendering, animatable human models, and handling casual videos. Each approach is summarized with its key contributions and evaluated based on its capabilities related to these three aspects. 

\begin{table}[hbtp]
\caption{\label{tab:3}Comparison of approaches in terms of articulation}
\centering
\resizebox{\textwidth}{!}{\begin{tabular}{|l|l|l|l|l|}
\hline
Paper &
  Main contribution &
  \begin{tabular}[c]{@{}l@{}}Free-viewpoint \\ rendering\end{tabular} &
  \begin{tabular}[c]{@{}l@{}}Animatable \\ human models\end{tabular} &
  \begin{tabular}[c]{@{}l@{}}Casual \\ videos\end{tabular} \\ \hline
iNeRF \cite{iNeRF}          & Estimating pose from a single image using NeRF                 & No  & No  & No  \\ \hline
NARF \cite{NARF}            & Representing and rendering articulated objects using NeRF      & No  & Yes & No  \\ \hline
Animatable-NeRF \cite{Animatable} & Animating human bodies using NeRF                              & Yes & Yes & No  \\ \hline
HumanNeRF \cite{HumanNeRF}       & Free-viewpoint rendering of moving people from monocular video & Yes & Yes & No  \\ \hline
BANMo \cite{BANMo}           & Building animatable 3D neural models from many casual videos   & No  & Yes & Yes \\ \hline
\end{tabular}}
\end{table}

\section{Scene Editing}
NeRF has limitations in its ability to edit scenes, such as adding or removing objects, and changing lighting or materials. This is because NeRF learns to represent a scene as a dense field of radiance, which means that any changes to the scene would require modifications to the radiance field. This can be computationally expensive and time-consuming. Additionally, NeRF is trained on a dataset of images that have been captured in a specific way, so it may need to be able to generalize to scenes that have been captured in a different way.

Liu et al. \cite{edit} present a novel method named Edit Conditional Radiance Fields (ECRF) for editing 3D objects represented by conditional radiance fields (CRFs). The method enables users to change the appearance of a local part, modify the local shape, or transfer color or shape from a target object instance. The CRF is trained over a set of shapes belonging to a class, and each shape instance is represented by latent shape and appearance vectors. The proposed model includes a shared shape network, an instance-specific shape network, a fusion shape network, and two separate networks for density and radiance predictions. 
The editing process is achieved by optimizing a loss function that balances accuracy and efficiency. The strategies involve updating specific network layers to speed up the optimization process and reduce the computational cost, subsampling user constraints during training, and feature caching to avoid unnecessary computation during optimization. These strategies achieve both accuracy and efficiency in editing the instance.
The authors demonstrate the proposed method's effectiveness on various 3D objects, including chairs, cars, and animals. The results show that the method is able to accurately and efficiently edit the appearance and shape of 3D objects represented by CRFs.
The main limitations of the method are that it can take over a minute for a user to get feedback on their shape edit, and the method fails to reconstruct novel object instances that differ from other class instances. However, the authors are optimistic that NeRF rendering time improvements will help to reduce the interactive time and that future work could address the limitations of the method.

\begin{figure}[htbp]
\centering
\includegraphics[width=\linewidth]{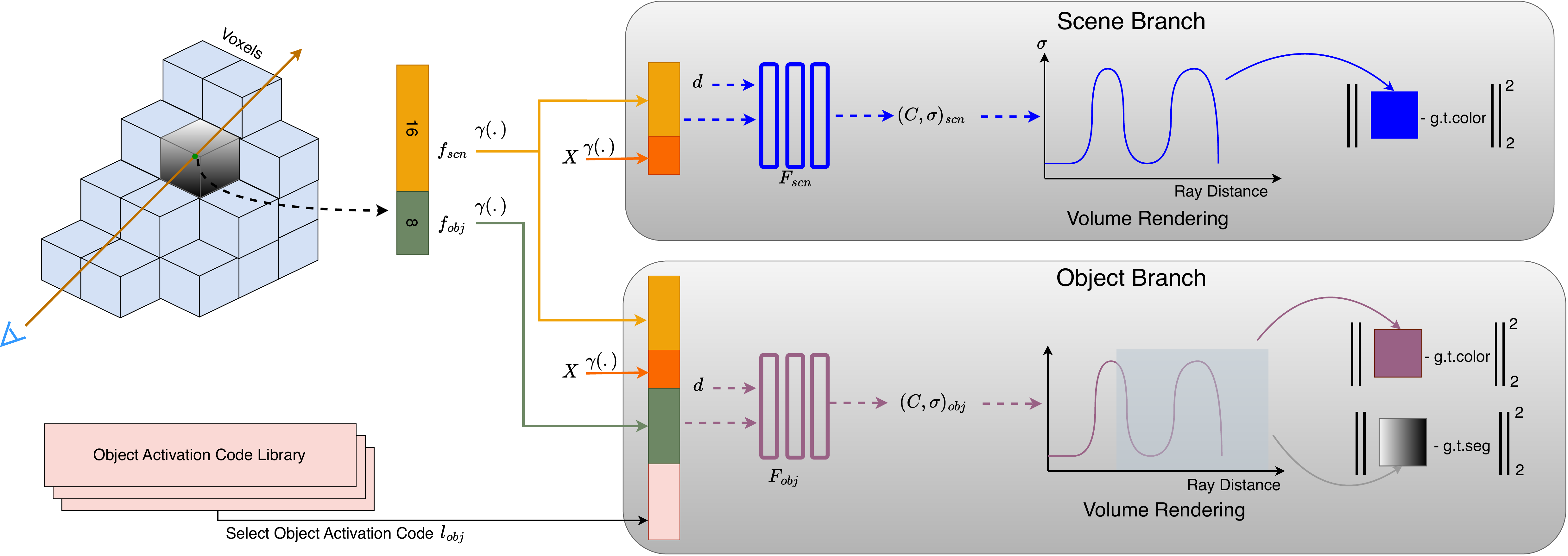}
\caption{LOCNeRF uses a two-pathway neural network architecture for learning an object-compositional neural radiance field. The first pathway is the scene branch, which takes as input the 3D spatial coordinate $X$, voxel features $F_{scn}$ interpolated at $X$, and the ray direction $d$. It outputs the predicted color $C_{scn}$ and density $\sigma_{scn}$ for the full scene. The second pathway is the object branch. In addition to $X$, $F_{scn}$, and $d$, it also takes as input voxel features $F_{obj}$ shared across objects and an object activation code $l_{obj}$ that identifies each object. The object branch is trained to output the color $C_{obj}$ and density $\sigma_{obj}$ for only the target object identified by $l_{obj}$, with everything else empty. At test time, LOCNeRF can render each object individually by passing its activation code to the object branch.
}
\label{fig:locnerf}
\end{figure}

ECRF can only edit the color, shape, and appearance of a single object. On the other hand, LOCNeRF \cite{LearningObject} can edit scenes of multiple objects and remove and add objects. This is because LOCNeRF learns a separate latent representation for each object in the scene, while ECRF is trained on a single object. LOCNeRF uses a two-pathway architecture (Figure \ref{fig:locnerf}) consisting of the scene and object branches. The scene branch encodes the scene geometry and appearance, while the object branch encodes each individual object in the scene.
The object branch takes in the embedded object voxel feature and object activation code to output the color and opacity for the desired object. Object activation codes are learnable code vectors that allow the object branch of the neural network to specialize in representing and rendering specific object instances, enabling object-level editing capabilities in the neural rendering framework. The object activation code is unique to each object and acts as a condition for the object branch to produce the desired output while everything else remains empty.
The method is jointly optimized for the scene branch and the object branch, with the total loss defined as the sum of the loss of the two branches. The editable scene rendering pipeline is divided into the background and the object stages.
In the background stage, the scene color and opacity are obtained while pruning the point sampling at the target region. In the object stage, rays are shot at the target objects, and the object-specific color and opacity are transformed to the desired location. Finally, all opacity and colors are aggregated, and pixel colors are rendered with quadrature rules \cite{max1995optical}. Quadrature rules provide a way to numerically estimate integrals, such as integrating color and opacity along a ray for volume rendering. LOCNeRF leverages standard quadrature rule integration techniques to differentiably render the object-compositional neural radiance fields for both the scene and the object branches.
The proposed method is the first editable neural scene rendering method that supports high-quality novel view rendering and object manipulation. It shows competitive performance for static scene novel-view synthesis and object-level editing. The method can achieve realistic rendering and enable high-level editing tasks, such as moving or adding objects.

NeRFs have been successfully used for various tasks, including scene editing, novel view synthesis, and 3D object reconstruction. However, one remaining challenge is decoupling dynamic and static objects from a monocular video.
D$^2$NeRF \cite{D2NeRF} from a Monocular Video presents a novel approach to this problem. The proposed method learns two separate NeRFs for the static background and the dynamic objects. The static NeRF is trained using standard NeRF techniques, while the dynamic NeRF is trained using a modified version of HyperNeRF \cite{park2021hypernerf}  that is better suited for non-rigid motion.
The two NeRFs are then combined to reconstruct the scene. The authors propose a novel loss function that encourages the two NeRFs to be consistent with each other while also allowing dynamic objects to move and change over time.
The authors evaluate their method on a new dataset containing various dynamic objects and shadows. They show that their method outperforms existing approaches in terms of both decoupling accuracy and 3D scene reconstruction quality.
The proposed method is a promising new approach to decoupling dynamic and static objects from a monocular video. It is self-supervised, meaning it does not require any ground-truth annotations. This makes it more scalable and applicable to real-world scenarios.
However, the proposed method does have some limitations. For example, it relies on precise camera registration for successful decoupling and reconstruction. Additionally, view-dependent radiance changes caused by reflective surfaces may be misinterpreted as dynamic effects, leading to incorrect decoupling.

Unlike D$^2$NeRF, that can only decouple dynamic from static objects, DFF \cite{dff} can decompose scene-specific NeRFs into arbitrary semantic units using text and image queries. This allows for versatile scene edits without the need for retraining. DFFs extend NeRFs by adding decoders for other quantities of interest, such as semantic labels, to compute density and view-dependent color. The authors propose a 3D zero-shot segmentation approach that uses open-set text labels or other feature queries to create a feature branch that models a 3D feature field describing the semantics of each spatial point. The authors propose to supervise the feature field using a pre-trained pixel-level image encoder as a teacher network and optimize it by minimizing the difference between rendered features and the teacher's features. To maintain the quality of reconstructed geometry, a stop-gradient is applied to density in rendering features.
The DFF model calculates the probability of an object label for a point in 3D space, and this segmentation can be used for interactive editing without retraining. This method allows for blending two NeRF scenes based on a segmentation field, and various edits can be applied to specific regions using this segmentation. The method also enables complex edits, including optimization-based methods, and allows for selective editing of particular regions in multi-object scenes.
However, the DFF framework has two limitations. Firstly, the performance of the distillation process is limited by the teacher model and the resolution of the teacher encoders. This limitation may result from coarse-grained DFFs that cannot understand specific text queries. Secondly, DFF relies on 3D reconstruction by NeRF, which can result in geometry errors in radiance fields that make the supervision of DFFs noisy.

Table \ref{tab:4} offers a comprehensive comparison of various approaches in the domain of scene editing, focusing on their methods, types of edits facilitated, and their respective limitations. This table succinctly presents the features and limitations of each approach, aiding in assessing their suitability for various scene editing tasks. 

\begin{table}[hbtp]
\caption{\label{tab:4}Comparison of approaches in terms of scene editing}
\centering
\resizebox{\textwidth}{!}{\begin{tabular}{|l|l|l|l|l|}
\hline
Paper &
  Method &
  Edits &
  Limitations \\ \hline
ECRF \cite{edit} &
  Conditional radiance fields (CRFs) &
  \begin{tabular}[c]{@{}l@{}}Edit color, shape, and appearance\\ of a single object\end{tabular} &
  \begin{tabular}[c]{@{}l@{}}Can take over a minute for \\ feedback; fails to reconstruct \\ novel object instances\end{tabular} \\ \hline
LOCNeRF \cite{LearningObject} &
  \begin{tabular}[c]{@{}l@{}}Neural scene rendering with object\\editing capabilities\end{tabular} &
  \begin{tabular}[c]{@{}l@{}}Edit multiple objects as well as\\remove and add objects\end{tabular} &
  Requires precise camera registration \\ \hline
D$^2$NeRF \cite{D2NeRF} &
  \begin{tabular}[c]{@{}l@{}}Decoupling of dynamic and static objects \\ from a monocular video\end{tabular} &
  \begin{tabular}[c]{@{}l@{}}Edit dynamic and \\static objects independently\end{tabular} &
  \begin{tabular}[c]{@{}l@{}}Relies on precise camera registration, \\ may misinterpret view-dependent \\ radiance changes\end{tabular} \\ \hline
DFFs \cite{dff} &
  \begin{tabular}[c]{@{}l@{}}Distilled feature fields (DFFs) for local \\ and interactive editing of NeRFs\end{tabular} &
  \begin{tabular}[c]{@{}l@{}}Edit appearance, add, remove, \\ and move objects\end{tabular} &
  \begin{tabular}[c]{@{}l@{}}Performance limited by the teacher model \\ and the resolution of teacher encoders; \\ relies on 3D reconstruction by NeRF\end{tabular} \\ \hline
\end{tabular}}
\end{table}

\section{Discussion}
The NeRF approach has demonstrated impressive results in both the quality and efficiency of rendering photorealistic images of 3D scenes. Its use of a neural network representation allows for capturing complex, non-linear relationships in the data, resulting in a more accurate and detailed representation of the scene. This implicit representation also allows for more efficient rendering processes, eliminating the need for pre-processing steps such as voxelization or meshing. One of the most notable advantages of the NeRF approach is its ability to produce high-quality images of novel views of a scene, even under challenging lighting conditions. This capability is particularly valuable in applications such as virtual reality, robotics, and autonomous vehicles, where synthesizing views from arbitrary viewpoints is critical. In addition, NeRF is able to capture complex light transport effects, such as reflections and refractions, that are challenging for other methods to reproduce. Combined with its ability to handle variations in input data, NeRF has emerged as a promising method for view synthesis and other computer vision applications. Finally, the success of NeRF has inspired further research and development in the field of implicit representation learning and view synthesis, which may lead to further breakthroughs in computer vision and graphics. 

Table \ref{tab:5} presents a comparison of various methods for image reconstruction using various evaluation metrics such as PSNR (Peak Signal-to-Noise Ratio), SSIM (Structural Similarity Index), and LPIPS (Learned Perceptual Image Patch Similarity). A higher PSNR indicates a more faithful reconstruction. SSIM is a measure of the similarity between two images and is calculated based on the local statistics of the images. A higher SSIM indicates a more similar image pair. LPIPS is a measure of the perceptual similarity between two images. It is calculated using a deep neural network that has been trained to distinguish between real and fake images. A lower LPIPS indicates a more perceptually similar image pair.

The PSNR metric shows that most models perform poorly on the DTU dataset, with NeRF having the lowest score of 8.00, while on some of the other datasets, models such as PointNeRF, Mip-NeRF, and HumanNeRF perform much better with scores of 33.31, 33.09, and 36.01, respectively.

The SSIM metric also shows similar results, with NeRF performing the worst on the DTU dataset, while PointNeRF, NuroFusion, and HumanNeRF perform much better on their respective datasets.

The LPIPS metric shows that most models have high values, indicating room for improvement in the neural rendering methods. However, some models, such as KiloNeRF, Edit-NeRF, and NeRF-Editing, have relatively lower LPIPS scores, indicating that they perform better in perceptual similarity.  

\begin{table}[ht]
\caption{\label{tab:5}Comparison of various NeRF methods evaluated on different datasets based on PSNR, SSIM, and LPIPS scores}
\resizebox{0.7\textwidth}{!}{\begin{tabular}{|l|c|c|c|l|}
\hline
Paper          & PSNR    & SSIM   & LPIPS  & Dataset Used            \\ \hline
NeRF \cite{1}           & 8.00    & 0.286  & 0.703  & DTU \cite{jensen2014large}                     \\ \hline
CoCo-INR \cite{cocoinr}        & 26.738  & 0.852  & 0.298  & DTU \cite{jensen2014large}                      \\ \hline
DietNeRF \cite{Diet}      & 14.242  & 0.481  & 0.487  & DTU \cite{jensen2014large}                      \\ \hline
PointNeRF  \cite{pointNerf}    & 33.31   & 0.978  & 0.049  & NeRF Synthetics \cite{1}           \\ \hline
NuroFusion \cite{NeRFusion}     & 31.25   & 0.953  & 0.069  & NeRF Synthetics \cite{1}           \\ \hline
FastNerf  \cite{FastNeRF}     & 29.155  & 0.936  & 0.053  & NeRF Synthetics \cite{1}            \\ \hline
KiloNeRF \cite{KiloNeRF}      & 31.00   & 0.95   & 0.03   & NeRF Synthetics \cite{1}            \\ \hline
SteerNeRF \cite{SteerNeRF}      & 30.97   & 0.948  & 0.065  & NeRF Synthetics \cite{1}           \\ \hline
MobileNeRF \cite{MobileNeRF}     & 30.90   & 0.947  & 0.062  & Syntatic 360 \cite{1}           \\ \hline
Mip-NeRF \cite{mipNerf}       & 33.09   & 0.961  & 0.043  & Blander \cite{mipNerf}                 \\ \hline
Mega-NeRF \cite{MegaNeRF}      & 22.08   & 0.628  & 0.489  & UrbanScene3d \cite{lin2022capturing}          \\ \hline
Pix2NeRF \cite{Pix2NeRF}       & 18.14   & 0.84   & -      & ShapeNet-SRN \cite{sitzmann2019scene}            \\ \hline
Block-NeRF \cite{blocknerf}          & 23.60   & 0.649  & 0.0417 & Alamo Square dataset    \\ \hline
LOLNeRF \cite{Lolnerf}        & 25.3    & 0.836  & 0.491  & CelebA-HQ \cite{karras2018progressive}             \\ \hline
FDNeRF \cite{FDNeRF}        & 24.847  & 0.821  & 0.142  & VoxCelebdataset \cite{NagraniCZ17}         \\ \hline
ECRF \cite{edit}         & 37.67   &   -     & 0.022  & PhotoShapes \cite{3272127}             \\ \hline
NeRF-Editing \cite{NeRF-Editing}   & 29.62   & 0.975  & 0.024  & Mixamo \cite{Mixamo59}    \\ \hline
D$^2$NeRF \cite{D2NeRF}        & 34.14   & 0.979  & 0.090  & Bag                     \\ \hline
DFFs \cite{dff}           & 32.85   & 0.932  & 0.162  & Replica dataset          \\ \hline
LOCNeRF \cite{LearningObject} & 15.0607 & 0.585  & 0.522  & ToyDesk                 \\ \hline
NARF \cite{NARF}           & 30.86   & 0.9586 & -      & THUman                  \\ \hline
HumanNeRF \cite{HumanNeRF}     & 36.01   & 0.9897 & 0.0356 & Multi-view dataset\cite{HumanNeRF} \\ \hline
\end{tabular}}
\end{table}

NeRF has become a popular tool for synthesizing photorealistic 3D scenes from 2D images. However, some challenges and limitations still remain. One of the major drawbacks of NeRF is its memory requirements, which can be a significant bottleneck for practical applications. The high memory consumption of NeRF is due to the need to store the neural network parameters, which can be challenging for large-scale scenes or high-resolution images. The computational expense of NeRF is also a limitation, particularly during training. Evaluating the neural network for each ray in the scene is time-consuming, and techniques proposed to accelerate the training of NeRF may come at the cost of decreased performance. Another limitation of NeRF is its inability to handle dynamic scenes or moving objects. NeRF only learns a single representation of the scene, which needs to be improved to address changes in the scene over time. Attempts to extend NeRF to handle dynamic scenes are an active research area. In addition, NeRF is sensitive to the quality of the input images, as it relies on accurate depth and camera pose estimates, which may be challenging to obtain in practice, especially for complex scenes. Finally, NeRF needs more interpretability, as there is a need to understand why particular views are synthesized correctly or incorrectly. Developing techniques for understanding and visualizing learned representations of NeRF is an important direction for future research. Furthermore, NeRF's slow inference speed, reliance on accurate pose estimation and multiple views, and limited effectiveness in scenarios with sparse views or poor camera calibration are additional limitations that must be addressed. Continued research efforts are needed to overcome these limitations and unlock NeRF's full potential.

\section{Conclusion}
The Neural Radiance Fields (NeRF) method has shown great potential for solving the challenging problem of image-based view synthesis. It provides a powerful and flexible representation of the 3D scene geometry and appearance using a continuous implicit function defined by a neural network. Our review has highlighted the various extensions and improvements to the original NeRF method, demonstrating the potential of this approach to solve more complex application scenarios. The extensions and improvements to the original NeRF method have made it more versatile, efficient, and capable of solving complex application scenarios. However, the high computational cost and lack of interpretability remain significant challenges that must be addressed in future research. Despite these limitations, NeRF has opened new possibilities for virtual and augmented reality, gaming, and robotics. To further advance the field, future research directions for NeRF include improving interpretability, reducing computational overhead to achieve real-time rendering, exploring new application scenarios, and enhancing the scalability and versatility of the method. The NeRF has undoubtedly paved the way for further research directions and possibilities, making it an exciting time for computer vision.

\bibliographystyle{unsrt}
\bibliography{subsections/0_main_tex}
\end{document}